\pdfoutput=1
\documentclass[10pt]{article}
\usepackage{lab_preprint_2}
\usepackage{adjustbox}
\usepackage{enumitem}
\usepackage{booktabs}
\usepackage{wrapfig}
\usepackage{graphicx}
\usepackage{caption}
\usepackage{makecell}
\usepackage{xcolor}
\usepackage{algorithm}
\usepackage{algpseudocode}
\usepackage{arydshln}
\usepackage[normalem]{ulem}
\usepackage{subcaption}

\usepackage{amsmath,amsfonts,bm}

\def\eqref#1{equation~\ref{#1}}

\def\1{\bm{1}}

\DeclareMathAlphabet{\mathsfit}{\encodingdefault}{\sfdefault}{m}{sl}
\SetMathAlphabet{\mathsfit}{bold}{\encodingdefault}{\sfdefault}{bx}{n}

\usepackage{hyperref}
\hypersetup{pdftitle={GPWM: Modeling the World as 3D Particle Motion},
  pdfauthor={Robert Ren, Matthew Bronars, He Zhu, Yikai Tang, Jianing Zhang, Sergey Zakharov, Pavel Tokmakov, Katerina Fragkiadaki}}
\usepackage{url}

\input{macros}

\title{\gpwmgradient{}: Modeling the World as 3D Particle Motion}

\author{Robert~Ren\textsuperscript{1,$\ast$}, Matthew~Bronars\textsuperscript{1,$\ast$}, He~Zhu\textsuperscript{1}, Yikai~Tang\textsuperscript{1}, Jianing~Zhang\textsuperscript{1},\\ Sergey~Zakharov\textsuperscript{2}, Pavel~Tokmakov\textsuperscript{2}, Katerina~Fragkiadaki\textsuperscript{1}}
\affiliations{\textsuperscript{1}Carnegie Mellon University, \textsuperscript{2}Toyota Research Institute}
\equalcontrib{Equal contribution, ordered by coin flip}
\metadata{%
  \metaline{Contact}{\mbox{\texttt{\{robertre, mbronars, katef\}@andrew.cmu.edu}}}
  \metaline{Project page}{\url{https://robertren1122.github.io/gpwm/}}
}
\logos{\lplogo[0.5in]{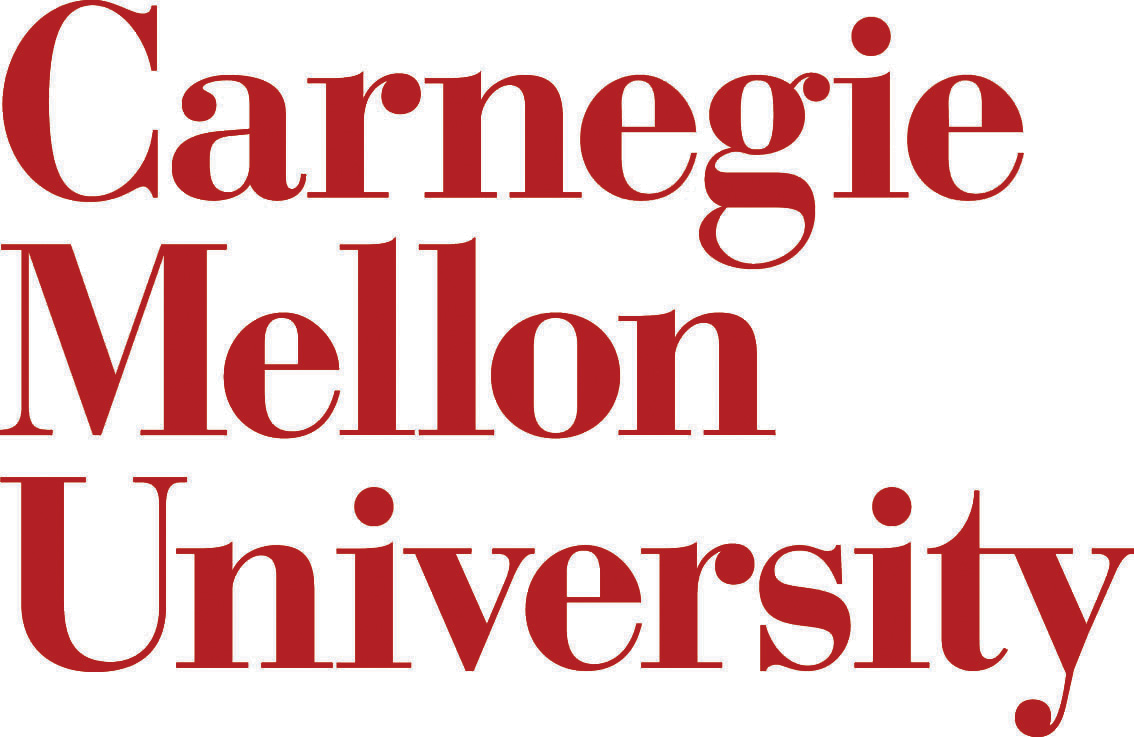}{CMU logo}\hspace{14pt}\lplogo[0.5in]{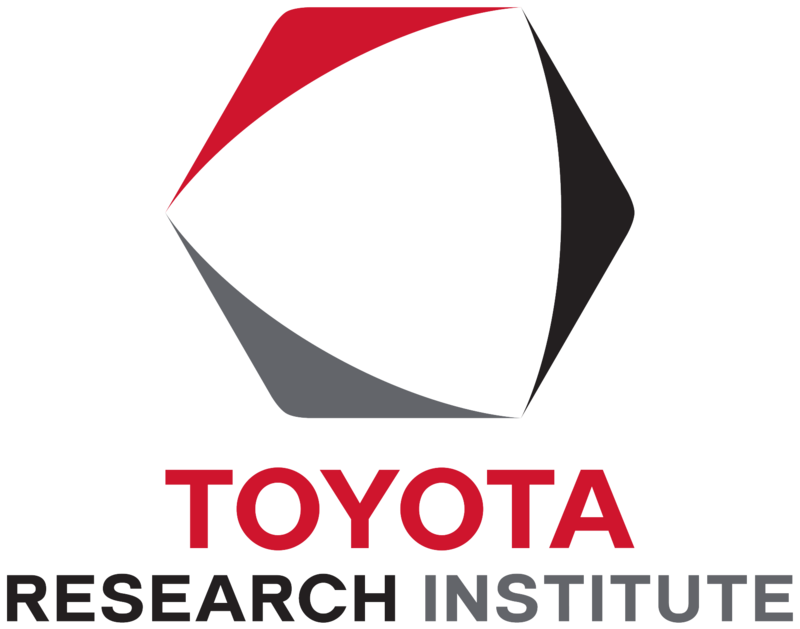}{TRI logo}}

\begin{document}

\maketitle

\newcommand{\website}[2]{\href{#1}{#2}}
\begin{abstract}

We introduce \gpwmgradient{}, a \textbf{G}enerative \textbf{P}article
\textbf{W}orld \textbf{M}odel that predicts how 3D scenes evolve by
generating the motion of particles and rigid bodies over time. Unlike video world models that generate scene
appearance, \model{} operates directly in 3D physical space. A single
representation and architecture cover rigid, deformable, cloth, and
granular materials, in both passive physics and robotic interaction.
\model{} is an object-aware multi-resolution transformer that combines
flow matching with diffusion forcing to generate coherent physical
trajectories far beyond its training horizon. We train and evaluate
\model{} on \dataset{}, a 25.5-million-frame corpus spanning diverse
geometries, materials, interactions, and robot embodiments. 
\model{} outperforms the best prior learned-dynamics baseline in each of six physical domains, reducing prediction error by 31\% on average.
Breaking from the domain-specific setting of prior methods, a single \emph{Unified} \model{} models these heterogeneous regimes with one set of weights, largely outperforming per-domain baselines and even surpassing our own specialist models in two domains.
\model{} further generalizes to unseen geometries, physical properties,
object counts, and rollout horizons. Beyond prediction, \model{} enables
contact-rich planning by learning a generative prior over joint
actuator--object motion. Instead of searching over unconstrained
actions, planning searches the model's latent noise for interactions
that achieve a desired object-space goal. With the same world model,
noise-space planning substantially outperforms action-space search on
pushing and picking, where complex contact sequences are hard to
discover. Executed open-loop in simulation, the resulting plans achieve
goal-directed manipulation without reference action trajectories or
goal-specific policies.

\end{abstract}

\providecommand{\dataset}{\textsc{ParticleWorlds}}
\begin{figure}[h!]
  \centering
  \includegraphics[width=\textwidth]{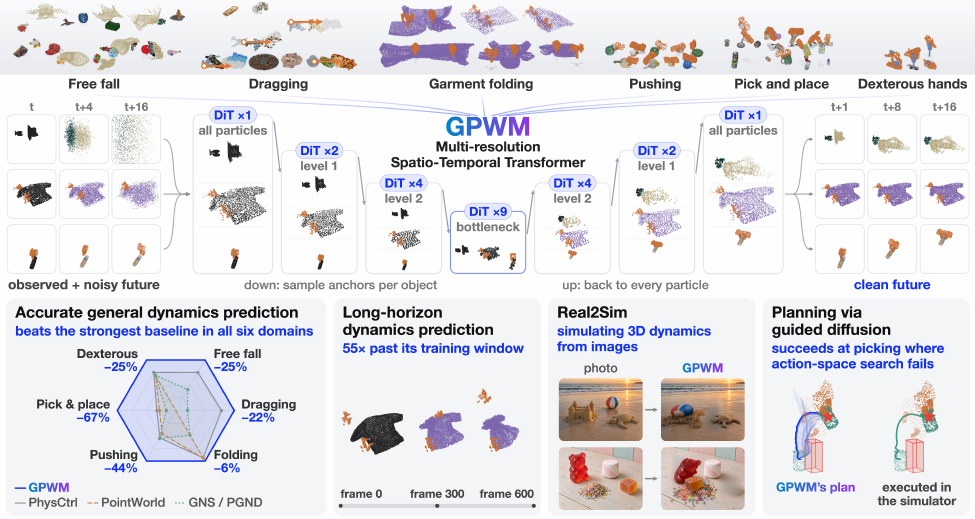}
  \relax
\caption{\textbf{\model{} learns 3D particle dynamics across six physical domains.}
Top: samples from the six domains of \dataset{}. Middle: given the observed frames and a noisy
future, a U-shaped multi-resolution spatio-temporal transformer denoises the future motion of
every particle. Bottom: \model{} (the better of its specialist and unified models in each domain)
lowers prediction error relative to the strongest baseline in every domain, by 31\% on average, rolls out
55$\times$ past its 16-frame training window, simulates 3D dynamics from images, and plans via
guided diffusion, succeeding at picking where action-space search fails.}
  \label{fig:teaser}
  \relax
\end{figure}

\relax
\section{Introduction}
\relax

A general-purpose physical world model should predict how a 3D scene
evolves and how to act to bring it toward a desired state. This requires
modeling diverse objects and materials, maintaining coherent dynamics
over long horizons, and reasoning through physical contact.
Recent generative video world models capture rich and multimodal scene
evolution~\citep{cosmos3,wan21,bruce2024genie,guo2026ctrl},
but generate visual appearance rather than explicitly representing the
motion and interactions of physical objects in 3D.
Particle- and point-based models instead represent physical state
directly in 3D~\citep{li2019learning,sanchez2020learning,pfaff2020learning,
huang2025particleformer3dpointcloud,duisterhof2026pointzero}. However, existing learned 3D
dynamics models are typically limited by deterministic prediction,
short generation horizons, or specialization to particular materials
and interactions~\citep{sanchez2020learning,shi2024robocraft,
zhang2025particlegrid,wang2026physctrl,chen2026physiformer}.
We ask: \emph{can a generative model of 3D particle motion capture
diverse physical dynamics, predict over long horizons, and support
planning through contact?}

We introduce \model{}, a generative world model that represents 3D
scenes as particles and rigid bodies and predicts their motion over
time (Figure~\ref{fig:teaser}). Deformable objects are modeled through
particle displacements and rigid bodies through body-level
transformations, providing a common representation for rigid,
deformable, cloth, and granular dynamics.
\model{} conditions on observed 3D scene history and, when available,
physical properties and future actuator motion. Actuators share the
same geometric representation as other scene components, allowing
their trajectories to either condition physical prediction or be
generated jointly with object motion for planning.

Learning generative particle dynamics at scale presents two challenges:
scenes may contain thousands of particles across interacting objects,
and prediction errors accumulate over long rollouts.
\model{} addresses the first with object-aware multi-resolution
tokenization and spatio-temporal attention in relative 3D coordinates.
To address the second, we combine flow matching with
\emph{diffusion forcing}, independently varying noise levels across
future frames. This enables \emph{streaming generation}, where
near-future states are progressively resolved and committed while
farther-future states remain uncertain and revisable, allowing
generation far beyond the training horizon.

We train and evaluate \model{} on \dataset{}, a corpus of 25.5 million
simulated frames spanning rigid, elastic, elastoplastic, granular,
and cloth dynamics, with diverse geometries, physical properties,
interactions, and robot embodiments.
Unlike prior learned dynamics models trained for individual physical
domains~\citep{sanchez2020learning,zhang2025particlegrid,
huang2026pointworld,wang2026physctrl}, we also train a single
\emph{Unified} \model{} across heterogeneous physical regimes.
Across six domains, the best \model{} variant reduces prediction
error by 31\% on average relative to the strongest applicable prior
baseline. Remarkably, the Unified model outperforms domain-specific
specialists on Dragging and Pick and Place, demonstrating that heterogeneous
physical dynamics can be learned within one set of weights.
\model{} further generalizes to unseen geometries, physical
properties, object counts, and long rollout horizons.

Beyond prediction, \model{} learns a \emph{generative prior over
physical interactions} that enables contact-rich planning.
Conventional action-space planning searches actuator commands and
predicts their consequences, but can struggle to discover complex
contact sequences or exploit world-model errors through
out-of-distribution actions. By jointly generating actuator and
object trajectories, \model{} instead allows planning in its latent
noise space, biasing search toward learned interaction patterns.
We optimize this noise using either gradient-based latent noise
optimization~\citep{eyring2024reno,ben2024d,
karunratanakul2024optimizing} or gradient-free sampling, guided by
sparse rewards on desired object configurations.
Using the same world model, noise-space planning substantially
outperforms action-space search on pushing and picking.
The resulting actuator trajectories execute successfully in
simulation without reference action trajectories or goal-specific
policies.

Our contributions are as follows.
\textbf{(1)} We introduce \textbf{\model{}}, a generative world model
that predicts the motion of particles and rigid bodies across
heterogeneous physical regimes, with streaming generation far
beyond its training horizon.
\textbf{(2)} We introduce \textbf{\dataset{}}, a 25.5M-frame corpus
spanning diverse physical dynamics, interactions, and robot embodiments.
\textbf{(3)} \model{} achieves \textbf{state-of-the-art physical prediction}
across diverse domains, generalizes to unseen geometries, physical
properties, object counts, and rollout horizons, and demonstrates
that a single Unified model can outperform domain-specific specialists.
\textbf{(4)} We demonstrate \textbf{contact-rich planning through
generative interaction priors}, showing that searching the learned
distribution of actuator--object trajectories is substantially more
effective than direct action-space optimization.

\relax
\section{Related Work}
\relax

\paragraph{Physics simulators and learned 3D dynamics.}
Physics engines and differentiable simulators use explicit physical models for
prediction and control~\citep{todorov2012mujoco,hu2020difftaichi,du2021diffpd}, while learned
simulators infer dynamics from trajectory data using particle and mesh
representations~\citep{li2019learning,sanchez2020learning,pfaff2020learning}.
RoboCraft~\citep{shi2024robocraft} and RoboCook~\citep{shi2023robocook} model
elastoplastic manipulation; ParticleFormer~\citep{huang2025particleformer3dpointcloud} models multi-object,
multi-material interactions; RigidFormer~\citep{dou2026rigidformer} targets
multi-object rigid dynamics; and Particle-Grid Neural Dynamics~\citep{zhang2025particlegrid} uses a hybrid particle-grid representation for
deformable dynamics. PointWorld~\citep{huang2026pointworld} represents scene
motion and robot actions as 3D point flows for forecasting and model-predictive
control. These learned 3D dynamics models are predominantly domain-specific
and deterministic. Deterministic prediction becomes increasingly restrictive
as physical diversity grows: contact and interaction uncertainty can induce
multiple plausible futures that must otherwise collapse to a single
trajectory. \model{} instead learns a generative distribution over dynamics
within a common architecture spanning rigid, deformable, cloth, and granular
systems.

\relax

\paragraph{Generative physical dynamics and long-horizon prediction.}
The closest works to \model{} are PhysCtrl~\citep{wang2026physctrl} and
PhysiFormer~\citep{chen2026physiformer}. PhysCtrl generates 3D point
trajectories conditioned on physical parameters and forces across rigid,
elastic, sand, and plasticine materials. PhysiFormer uses a diffusion
transformer with attention factorized across space, time, and objects to
generate multi-object rigid and elastic trajectories. Both are trained over fixed,
relatively short generation horizons and retain the full set of 3D points or
vertices throughout the network, limiting scalability to densely represented
scenes; PhysiFormer, for example, trains on scenes with at most 356 vertices.
Both are also trained on substantially smaller datasets than \dataset{}.
In contrast, \model{} progressively compresses particles through a
multi-resolution hierarchy, scaling to scenes with up to 12{,}000 particles,
and combines this spatial scalability with diffusion
forcing~\citep{chen2024diffusion} for stable auto-regressive generation far beyond the
training horizon.

\relax
\paragraph{Generative models for planning.}
Diffuser~\citep{Janner2022} and Decision Diffuser~\citep{Ajay2022}
formulate decision making as trajectory generation conditioned or guided
by task objectives. Subsequent work explores adaptive
planning~\citep{Liang2023}, hierarchical trajectory
generation~\citep{Li2023}, and joint state--action generation for
control~\citep{dloc}.
Classifier- and reward-guided sampling typically steer intermediate
noisy states during generation, while latent noise optimization (LNO),
including D-Flow~\citep{ben2024d} and Diffusion Noise
Optimization~\citep{karunratanakul2024optimizing}, optimizes the initial
noise by differentiating through the generative process.

\model{} extends these ideas to contact-rich planning in 3D physical
environments. By jointly generating actuator and object trajectories,
it learns a prior over physical interactions that can be searched
using either gradient-based LNO or sampling-based optimization.
Unlike direct action-space search, which can propose out-of-distribution
actions and exploit world-model errors, searching this generative prior
biases planning toward plausible interaction trajectories.
Our experiments show that noise-space planning substantially
outperforms action-space optimization even when both use the same
world model.

\relax
\section{Method}
\label{sec:method}
\relax

\begin{figure}[t]
  \relax
  \centering
  \includegraphics[width=0.75\textwidth]{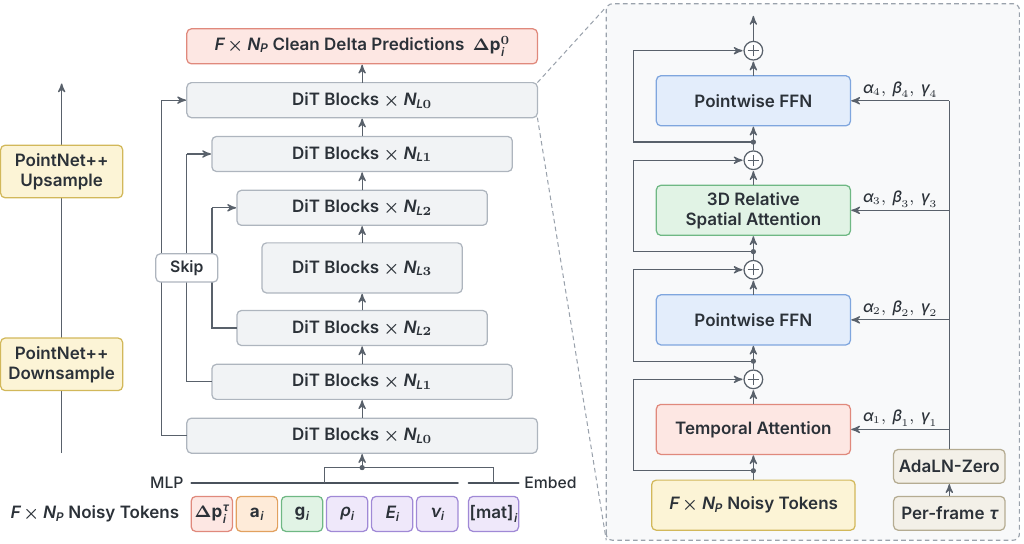}
  \relax
  \caption{\textbf{\model{} architecture.} Noisy per-particle tokens, which embed
  displacement, actuator motion, geometry, physical parameters, and material,
  pass through a multi-resolution DiT. PointNet++ layers pool particles into
  per-object anchors on the way down and interpolate back on the way up, and
  long skips connect matching levels. Each block (right) applies temporal
  attention, 3D relative spatial attention, and pointwise feed-forward layers,
  modulated by AdaLN-Zero from the per-frame noise level $\tau$, and the network
  outputs clean future displacements.}
  \label{fig:arch}
  \relax
\end{figure}

We formulate world modeling as conditional generation of 3D scene
trajectories. A scene contains $N$ tracked particles with positions
$\mathbf{P}_t\in\mathbb{R}^{N\times3}$ at frame $t$, grouped into objects,
including $K$ rigid bodies. We assume particle correspondences and object
assignments are known. Given $C$ observed frames $\mathbf{P}_{0:C-1}$,
\model{} models a distribution over the next $F$ frames, optionally conditioned
on material properties and prescribed actuator motion.

\model{} represents motion through particle displacements and rigid-body
transformations (Section~\ref{sec:representation}), processed by an object-aware,
multi-resolution spatio-temporal transformer (Section~\ref{sec:architecture}).
We train the model with flow matching and diffusion forcing
(Section~\ref{sec:training}) and use a streaming denoising schedule to generate
trajectories beyond the training window (Section~\ref{sec:inference}).
For planning, task-specific models jointly generate actuator and object
trajectories, allowing rewards on object outcomes to guide actuator motion
(Section~\ref{sec:planning}).

\relax
\subsection{Particle and Rigid-Body Representation}
\label{sec:representation}
\relax
\model{} represents scene motion through particle displacements and rigid-body
transformations. The generative variables reflect each component's physical
structure, while particle geometry provides a shared representation for
modeling interactions.
\relax
\paragraph{Particle representation.}
Let $N_p$ denote the number of particles represented by individual displacement
variables, with positions $\mathbf{P}_t^{p}\in\mathbb{R}^{N_p\times3}$.
Their frame-to-frame displacements are
  $  \Delta\mathbf{P}_t^{p}
    =
    \mathbf{P}_{t+1}^{p}-\mathbf{P}_t^{p}$,
and the clean future trajectory is $
    \mathbf{x}_0 =
    [\Delta\mathbf{P}_{C-1}^{p},\ldots,
     \Delta\mathbf{P}_{C+F-2}^{p}]
    \in\mathbb{R}^{F\times N_p\times3}. $
Future positions are recovered by accumulating these displacements from the
last observed frame. This makes the motion targets invariant to global
translation. Where available, we condition on material type and physical
parameters: Young's modulus $E$, Poisson's ratio $\nu$, and density $\rho$.
Section~\ref{sec:architecture} describes how motion, geometry, and conditioning
information are embedded into particle tokens.
\relax
\paragraph{Rigid-body representation.}
Each rigid body $k$ is represented by both a pose token and particle tokens. The pose token encodes its frame-to-frame transformation, $\Delta\mathbf{q}_k^t
    =
    [\Delta\mathbf{t}_k^t,\Delta\mathbf{r}_k^t]
    \in\mathbb{R}^{9},$
where $\Delta\mathbf{t}_k^t\in\mathbb{R}^{3}$ is a translation increment and
$\Delta\mathbf{r}_k^t\in\mathbb{R}^{6}$ represents the relative
rotation~\citep{zhou2019continuity}.
The clean future rigid-body trajectory is
$
    \mathbf{y}_0 =
    [\Delta\mathbf{q}^{C-1},\ldots,
     \Delta\mathbf{q}^{C+F-2}]
    \in\mathbb{R}^{F\times K\times9}.
$
We recover future poses from the last observed pose by accumulating
translations and composing the rotation matrices decoded from these
increments. Applying each pose to fixed body-frame particles preserves rigid shape by construction. These particles do not receive their own noise; they move with the noisy pose of their rigid body. Pose and particle trajectories are denoised jointly by the multi-resolution spatio-temporal transformer.
\relax
\paragraph{Actuator conditioning.}
We represent actuator commands using the same geometric variables as scene
motion: prescribed particle displacements $\mathbf{a}_i^t$ or rigid-body
transformations. Rods, grippers, and articulated hands are therefore described
through the motion of their physical components. When future actuator motion
is supplied, \model{} conditions on it to generate the resulting object
trajectories. For planning, task-specific models jointly generate actuator and
object trajectories, allowing rewards on future object configurations to guide
actuator motion (Section~\ref{sec:planning}).

\relax
\subsection{Multi-Resolution Spatio-Temporal Transformer}
\label{sec:architecture}
\relax
Direct attention over all particles and frames is prohibitively expensive for
scenes containing thousands of particles. We therefore use a multi-resolution
architecture that progressively compresses dense particles into a small set of
anchor tokens, performs most global interaction modeling at coarse resolution,
and propagates the resulting features back to the original particles (Figure~\ref{fig:arch}). 
\relax
\paragraph{Particle embeddings.}
Each particle $i$ at frame $t$ is embedded as
\[
\mathbf{h}_i^t =
\mathrm{MLP}\!\left(
[\Delta\mathbf{p}_i^t,\,
 \mathbf{g}_i^t,\,
 \boldsymbol{\phi}_i,\,
 \mathbf{a}_i^t]
\right)
+\mathbf{e}_{m_i},
\]
where $\Delta\mathbf{p}_i^t$ is its frame-to-frame displacement,
$\mathbf{g}_i^t$ encodes its displacement from its initial position together
with its absolute height,
$\boldsymbol{\phi}_i=[\log_{10}E_i,\nu_i,\rho_i]$ contains available physical
parameters, and $\mathbf{e}_{m_i}$ is a learned embedding of material type
$m_i$. For context frames, $\Delta\mathbf{p}_i^t$ is observed; for future
frames, it is the corresponding noisy generative variable.
\relax
\paragraph{Multi-resolution particle hierarchy.}
Particle tokens remain attached to physical particles at every resolution.
Coarse anchors are selected by farthest-point sampling (FPS) within each object
from the initial geometry. PointNet++-style layers~\citep{qi2017pointnet++}
transfer information between resolutions. During downsampling, each anchor
$a$ aggregates features from a neighborhood within the same object.
Suppressing feature projections, normalization, and noise-dependent gates,
this aggregation is written schematically as
\[
\tilde{\mathbf{h}}_a^t =
\max_{j\in\mathcal{N}(a,t)}
\mathrm{MLP}\!\left(
[\mathbf{h}_j^t,\,
 \mathbf{h}_j^t-\mathbf{h}_a^t,\,
 \mathbf{p}_j^t-\mathbf{p}_a^t]
\right),
\qquad
\mathbf{h}_a^t \leftarrow
\mathbf{h}_a^t+\tilde{\mathbf{h}}_a^t .
\]
During upsampling, coarse features are projected to the finer feature width
and interpolated back to finer particles using inverse-distance weighting,
\[
\mathbf{h}_i^t =
\sum_{a\in\mathcal{N}(i,t)}
w_{ia}\mathbf{h}_a^t,
\qquad
w_{ia} =
\frac{(\|\mathbf{p}_i^t-\mathbf{p}_a^t\|+\epsilon)^{-1}}
{\sum_{b\in\mathcal{N}(i,t)}
 (\|\mathbf{p}_i^t-\mathbf{p}_b^t\|+\epsilon)^{-1}} .
\]
Here $\epsilon>0$ handles coincident points. The interpolated features are
fused with the corresponding encoder features through a gated MLP.
Neighborhoods are recomputed at each frame from particle positions obtained by
integrating the current noisy displacement trajectory. Anchor identities
remain fixed throughout the temporal window while their positions evolve.
Feature width increases as particle token count decreases, and most
transformer blocks operate at coarse resolutions. Rigid-body pose tokens are
retained at every resolution and participate in attention alongside particle
tokens. Linear projections change their feature width between resolutions
without pooling them.
\relax
\paragraph{Factorized spatio-temporal attention.}
At each resolution, transformer blocks factorize attention into temporal and
spatial stages~\citep{arnab2021vivit}. Temporal attention operates along each
token trajectory within the denoising window, while spatial attention connects
tokens within each frame. At
full resolution, spatial attention is restricted within each object; at
coarser resolutions, it spans the entire scene, allowing the model to capture
interactions between objects and actuators. Each attention stage is followed
by a feed-forward network, and blocks are modulated with
adaLN-Zero~\citep{peebles2023scalable} using the frame-specific noise level.
Encoder and decoder stages are connected by Hunyuan-DiT-style long
skips~\citep{li2024hunyuandit}, preserving fine-scale information across the
multi-resolution hierarchy.
\relax
\paragraph{Relative 3D geometry.}
Particle inputs encode displacement from the initial position, and PointNet++
aggregation uses offsets relative to each anchor. Spatial attention applies
rotary position embeddings~\citep{heo2024rotary} to the 3D coordinates of
particle and pose tokens. Absolute height is retained to represent vertical
position relative to the ground plane.

\relax
\subsection{Flow Matching with Diffusion Forcing}
\label{sec:training}
\relax
We train \model{} with flow matching~\citep{lipman2022flow} and diffusion
forcing~\citep{chen2024diffusion}. Rather than assigning a single noise level
to the entire future trajectory, diffusion forcing samples an independent
noise level $\tau_t\in[0,1]$ for each future frame, collected in
$\boldsymbol{\tau}=(\tau_1,\ldots,\tau_F)$.
We normalize particle displacements and rigid-body increments before adding
noise. In this section, $\mathbf{x}_0$ and $\mathbf{y}_0$ denote these normalized
trajectories. For the particle and rigid-body trajectories, respectively,
\[
\mathbf{x}_{\boldsymbol{\tau}}^{t}
=
(1-\tau_t)\mathbf{x}_0^{t}
+
\tau_t\boldsymbol{\epsilon}_x^{t},
\qquad
\mathbf{y}_{\boldsymbol{\tau}}^{t}
=
(1-\tau_t)\mathbf{y}_0^{t}
+
\tau_t\boldsymbol{\epsilon}_y^{t},
\]
where
$\boldsymbol{\epsilon}_x^{t},\boldsymbol{\epsilon}_y^{t}\sim\mathcal{N}(0,I)$.
Thus different future frames can be presented to the model at different stages
of the generative process.

Similar to \citet{li2026back}, we use an endpoint parameterization: rather than
predicting the flow velocity directly, the network predicts the clean particle
and rigid-body trajectories,
\[
(\hat{\mathbf{x}}_0,\hat{\mathbf{y}}_0)
=
\mathcal{F}_{\theta}
\left(
\mathbf{x}_{\boldsymbol{\tau}},
\mathbf{y}_{\boldsymbol{\tau}},
\boldsymbol{\tau},
\mathbf{c}
\right),
\]
where $\mathbf{c}$ contains the observed context and any available physical
properties and action conditioning. The flow field used for sampling is
recovered from the noisy state and predicted clean endpoint. We do not reweight this objective in velocity space and instead directly optimize
\[
\mathcal{L}(\theta)
=
\mathbb{E}
\left[
\|\hat{\mathbf{x}}_0-\mathbf{x}_0\|_2^2
+
\gamma
\|\hat{\mathbf{y}}_0-\mathbf{y}_0\|_2^2
\right],
\]
where $\gamma$ balances the particle and rigid-body terms. No auxiliary
physics-based losses are used.

\relax
\subsection{Streaming Long-Horizon Generation}
\label{sec:inference}
\relax
Independent per-frame noise levels allow different future frames to occupy
different stages of denoising at inference time. Following diffusion
forcing~\citep{chen2024diffusion}, we use a staggered denoising schedule:
near-future frames are progressively resolved, while farther-future frames
remain at higher noise levels and continue to be revised. Once the earliest
frames are fully denoised, they are committed, the prediction window
advances, and fresh Gaussian noise is appended at the far end of the horizon.
Repeating this process allows \model{} to generate trajectories beyond the
$F$-frame training horizon while keeping the model's temporal window fixed.

\relax
\subsection{Planning with Generative Interaction Priors}
\label{sec:planning}

Planning uses task-specific \model{} models (Appendix~\ref{supp:planning_models}).
\model{} supports planning in either action space or the latent
space of its generative model. In action-space planning, candidate
actuator trajectories are evaluated using the action-conditioned
world model. Alternatively, by jointly generating actuator and
passive-object motion, \model{} learns a prior over physical
interactions that enables planning directly in latent noise space.
This biases search toward actuator--object trajectories represented
by the learned distribution, rather than unconstrained action
sequences.

We specify planning objectives through a differentiable reward on
the desired final configuration of passive objects. With the world
model frozen, we optimize the initial Gaussian noise through
\emph{latent noise optimization} (LNO)
~\citep{eyring2024reno,ben2024d,karunratanakul2024optimizing}.
Let $\mathbf{z}$ denote the initial noise for an $H$-frame rollout
and $\hat{\mathbf{s}}_{1:H}(\mathbf{z})$ the generated scene
trajectory, reconstructed in physical coordinates by denormalizing
and integrating predicted motion. With $\Pi_{\mathrm{pass}}$
selecting the passive-object configuration, LNO performs

\[
\mathbf{z}
\leftarrow
\mathbf{z}
+
\eta\nabla_{\mathbf{z}}
r\!\left(
\Pi_{\mathrm{pass}}
(\hat{\mathbf{s}}_H(\mathbf{z}))
\right),
\]

where $\eta$ is the guidance step size.
As a gradient-free alternative, we also apply the cross-entropy
method (CEM)~\citep{rubinstein1999cross} in the same latent noise space, iteratively sampling
and selecting noise vectors according to their predicted terminal
rewards.

For horizons $H>F$, we use the streaming generation procedure
of Section~\ref{sec:inference}. In LNO, gradients propagate through
successive generation windows, allowing terminal object-space
rewards to guide earlier actuator motion. The actuator component
of the optimized joint trajectory defines the resulting plan,
which is executed open-loop in simulation
(Section~\ref{sec:exp_guidance_proposed}).
Further details are provided in Appendix~\ref{supp:noise_opt}.

\relax
\section{\dataset{} Dataset}
\label{sec:dataset_proposed}
\relax

We introduce \dataset{}, a dataset of 25.5 million simulated training frames
spanning rigid objects, elastic and elastoplastic solids, granular materials,
and cloth. Its six domains cover passive dynamics and robotic
manipulation with rods, parallel-jaw grippers, and articulated hands
(Appendix~\ref{app:particleworlds}). Scenes range from isolated objects to cluttered
environments, with representations containing 256--12{,}000 particles and
geometries including irregular objects and garments from ten categories. This
breadth supports both domain-specific models and a unified world model
trained across materials and robot embodiments.

We generate trajectories for falling objects, deformable dragging,
garment manipulation, dexterous manipulation, rod and gripper pushing, and
pick-and-place~\citep{zhu2026skillweaver}. To combine
these sources, we convert simulator outputs into the common particle
representation of Section~\ref{sec:representation}, preserving particle
identities and object membership over time. Deformable bodies use tracked
material or mesh points; rigid bodies and actuator links use fixed surface
samples transformed by their recorded poses. Dataset sources, generation
procedures, and split definitions are detailed in
Appendix~\ref{app:particleworlds}.

\relax
\section{Experiments}
\label{sec:exp_proposed}
\relax

We evaluate \model{} along three dimensions:
\textbf{(1) prediction} across rigid, deformable, granular, and cloth dynamics;
\textbf{(2) generalization} across geometry, physical properties, scene
complexity, and rollout horizon, including joint training across heterogeneous
regimes; and \textbf{(3) contact-rich planning} from sparse object-space
rewards.

\relax
\paragraph{Baselines.}
We compare against four learned 3D dynamics models:
GNS~\citep{sanchez2020learning}, a graph-based particle simulator;
PGND~\citep{zhang2025particlegrid}, a hybrid particle--grid model;
PointWorld~\citep{huang2026pointworld}, a point-cloud world model; and
PhysCtrl~\citep{wang2026physctrl}, a generative model of physical dynamics.
Each is evaluated where its conditioning applies: PhysCtrl on all domains,
GNS on non-action-conditioned domains, PointWorld on robot-actuated domains, and PGND on
robot-actuated domains and Dragging. We report domain-specific \model{} models (\textbf{Specialist}) for
direct comparison and a single \textbf{Unified} model trained across all
domains.

\relax
\paragraph{Experimental setup.}
Unless stated otherwise, \model{} observes four context frames and predicts
16 future frames per window. Predictions are rolled forward to the end of each
trajectory, up to 898 frames for Garment Folding (over $55\times$ the
prediction window). Recorded actuator motion is provided for action-conditioned tasks;
passive tasks require no action input. All methods use the same evaluation
trajectories and available actuator conditioning.

\relax
\paragraph{Metrics.}
For physical prediction, we report mean Euclidean position error
(mm, $\downarrow$) over passive-object particles, averaging over particles and
frames and then equally across trajectories. We evaluate one sample from
\model{}. Because physical interactions can be multimodal, error against a
single recorded future does not fully capture physical plausibility; we
therefore provide extensive qualitative rollouts on our
\href{https://robertren1122.github.io/gpwm/}{project page}. For cloth, we report symmetric Chamfer distance
($\downarrow$). Planning is instead evaluated by task progress toward the goal
(Section~\ref{sec:exp_guidance_proposed}).

\relax
\subsection{Prediction across Physical Regimes}
\label{sec:exp_prediction_proposed}
\relax

Table~\ref{tab:prediction_overview_proposed} compares prediction across six
physical settings spanning rigid, deformable, granular, and cloth dynamics.

\relax
\paragraph{State-of-the-art prediction across diverse physics.} 
\model{} outperforms the strongest applicable baseline in all six settings,
reducing error by \textbf{6--67\%} (31\% on average) with the better of its
specialist and unified models (Table \ref{tab:prediction_overview_proposed}).
The gains span qualitatively different dynamics: error decreases from 53.0 to 40.0\,mm on Free Fall, 7.8 to 6.1\,mm on Dragging, and 23.9 to 13.3\,mm on Pushing.
Together, these results show that the same generative particle formulation
achieves strong prediction accuracy across substantially different physical
regimes.

\providecommand{\pendingresult}{\textemdash}

\begin{table}[!ht]
  \centering
  \caption{\textbf{Prediction across diverse physical settings.}
Prediction error in millimeters ($\downarrow$). Baselines and Ours
(Specialist) are trained separately for each setting; Ours (Unified) uses a
single model across heterogeneous settings. Graph/grid-based denotes GNS or
PGND, depending on the setting. --: not applicable. $^\dagger$Garment Folding reports symmetric Chamfer distance averaged over 898 predicted frames on 20 test trajectories.}
  \label{tab:prediction_overview_proposed}
  \small
  \setlength{\tabcolsep}{3pt}
  \renewcommand{\arraystretch}{1.2}
  \begin{tabular*}{\linewidth}{@{\extracolsep{\fill}}l *{6}{c}@{}}
    \toprule
    Method & \shortstack{Free\\Fall} & Dragging & \shortstack{Garment\\Folding$^\dagger$} & Pushing & \shortstack{Pick and\\Place} & \shortstack{Dexterous\\Hands} \\
    \midrule
    Graph/grid-based & 97.2 & 30.4 & 20.5 & 23.9 & 68.2 & 26.7 \\
    PointWorld & \pendingresult & \pendingresult & \underline{10.6} & 28.7 & 39.8 & 27.4 \\
    PhysCtrl & 53.0 & 7.8 & 10.7 & 24.5 & 34.1 & 24.8 \\
    \midrule
    Ours (Specialist) & \textbf{40.0} & \underline{6.5} & \textbf{10.0} & \textbf{13.3} & \underline{12.6} & \textbf{18.7} \\
    Ours (Unified) & \underline{41.5} & \textbf{6.1} & 10.8 & \underline{18.0} & \textbf{11.1} & \underline{22.6} \\
    \bottomrule
  \end{tabular*}
\end{table}
\relax
\paragraph{One model across heterogeneous physics.}
A single Unified \model{} spans heterogeneous materials, interactions, and
actuator embodiments within one set of weights. It outperforms all evaluated
prior methods on five of the six settings, and achieves
the best overall performance on Dragging and Pick and Place, improving even over the
\model{} specialists (Table \ref{tab:prediction_overview_proposed}). Domain-specific specialists remain
more accurate on the other settings, revealing a specialization gap while
demonstrating that substantially different physical regimes can be captured
within a single model.

\relax
\subsection{Generalization across Geometry and Physics}
\label{sec:exp_generalization_proposed}

\begin{figure}[t]
  \centering
  \includegraphics[width=\linewidth]{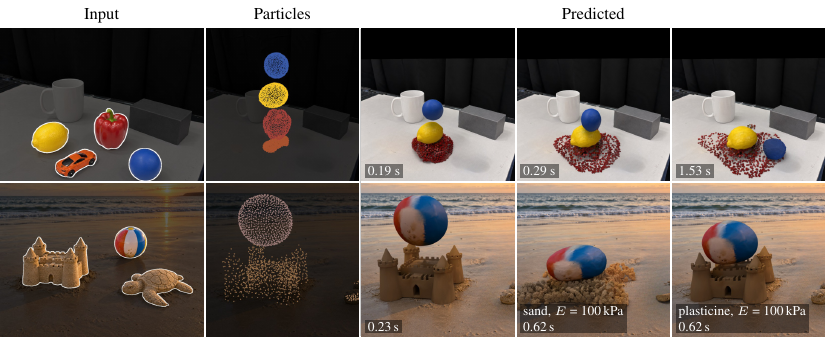}
  \caption{\textbf{Predictions from a single image.} Each row shows the input
  with the reconstructed objects outlined, their particles, and the predicted
  motion. Top, a tower of a toy car (elastic, 1\,MPa), a pepper (sand,
  100\,kPa), a lemon (plasticine, 100\,kPa), and a ball (elastic, 100\,kPa).
  Bottom, an elastic ball (10\,kPa) dropped onto a sand or a plasticine castle
  (100\,kPa).}
  \label{fig:drop_from_image}
\end{figure}

We conduct controlled generalization experiments in the Free Fall setting,
where geometry, object count, and material properties can be varied
independently. Table~\ref{tab:generalization_proposed} evaluates \model{} as
each factor moves beyond the training distribution. \model{} achieves the
lowest prediction error across all generalization conditions.
\relax
\paragraph{\model{} generalizes to unseen shapes and larger scenes.}
Free Fall training scenes contain one to four objects. At test time, we
introduce unseen geometries and increase scene complexity to five to seven
objects, reaching 7{,}168 particles. Unseen shapes increase specialist error
by less than 7\%, while combining unseen shapes with larger scenes raises
error only from 37.5 to 45.2\,mm. In contrast, PhysCtrl increases from
48.5 to 120.9\,mm. The Unified model is also stable across these shifts and ranks second in all four shape and object-count conditions.
\begin{table}[!ht]
  \centering
  \caption{\textbf{Generalization on Free Fall.} Mean particle position error
  in millimeters ($\downarrow$). Training uses one to four objects per scene.
  Stiffness values are seen (10, 100, 1{,}000\,kPa) or unseen within the training
  range (30, 300\,kPa). Material types are weighted equally.}
  \label{tab:generalization_proposed}
  \small
  \setlength{\tabcolsep}{7pt}
  \renewcommand{\arraystretch}{1.15}
  \begin{tabular*}{\linewidth}{@{\extracolsep{\fill}}l *{6}{c}@{}}
    \toprule
    & \multicolumn{2}{c}{1--4 objects} & \multicolumn{2}{c}{5--7 objects} & \multicolumn{2}{c}{Stiffness} \\
    \cmidrule(lr){2-3} \cmidrule(lr){4-5} \cmidrule(lr){6-7}
    Method & Seen shapes & Unseen shapes & Seen shapes & Unseen shapes & Seen & Unseen \\
    \midrule
    GNS & 96.8 & 97.2 & 100.9 & 97.6 & 116.7 & 113.2 \\
    PhysCtrl & 48.5 & 53.0 & 122.7 & 120.9 & \underline{51.2} & \underline{51.7} \\
    \midrule
    Ours (Specialist) & \textbf{37.5} & \textbf{40.0} & \textbf{45.1} & \textbf{45.2} & \textbf{47.3} & \textbf{48.9} \\
    Ours (Unified) & \underline{39.0} & \underline{41.5} & \underline{47.8} & \underline{47.5} & 54.5 & 53.1 \\
    \bottomrule
  \end{tabular*}
\end{table}
\relax
\paragraph{\model{} generalizes to unseen material stiffness.}
We next vary Young's modulus from 10\,kPa to 1\,MPa, covering values seen during
training and unseen values within the training range.
Models receive the material parameters and initial observations for each
simulation. \model{}'s error rises by less than 2\,mm on unseen values (47.3 to
48.9\,mm) while remaining below the strongest baseline. This demonstrates robust
prediction under interpolation of material stiffness.
\relax
\paragraph{\model{} generalizes to objects reconstructed from images.}
We also test \model{} on previously unseen object geometries reconstructed
from single RGB images (Figure~\ref{fig:drop_from_image}). SAM~3~\citep{sam3}
segments the objects, SAM~3D~\citep{sam3d} reconstructs their meshes, and
MoGe-2~\citep{moge2} estimates their metric scene placement. We sample surface
particles from the reconstructed meshes, assign physical properties, and use
the Free Fall specialist to predict their dynamics. The resulting trajectories
are rendered back into the scene, demonstrating that \model{} can operate
on reconstructed real-world geometry rather than only simulator-native assets.

\relax

\subsection{Planning with Generative Interaction Priors}
\label{sec:exp_guidance_proposed}

\model{} supports two approaches to model-based planning.
In \emph{action-space planning}, candidate actuator trajectories are evaluated
using an action-conditioned world model. However, searching unconstrained
action sequences presents two challenges: complex contact-rich behaviors
are difficult to discover without a suitable action proposal policy, and
optimization may exploit prediction errors by selecting actions outside
the model's training distribution.

\model{} offers an alternative: by jointly generating actuator and
passive-object trajectories, it learns a \emph{generative prior over physical
interactions}. Planning can then search the model's latent noise rather
than raw actuator commands, biasing optimization toward plausible
actuator--object trajectories. We optimize this noise using either
gradient-based latent noise optimization (LNO)
~\citep{eyring2024reno,ben2024d,karunratanakul2024optimizing}
or gradient-free sampling. In LNO, a differentiable terminal reward on the
passive-object configuration is backpropagated through the generated
trajectory to the initial noise (Sec.~\ref{sec:planning}).

\begin{wrapfigure}{R}{0.38\textwidth}
  \relax
  \centering
  \includegraphics[width=0.38\textwidth]{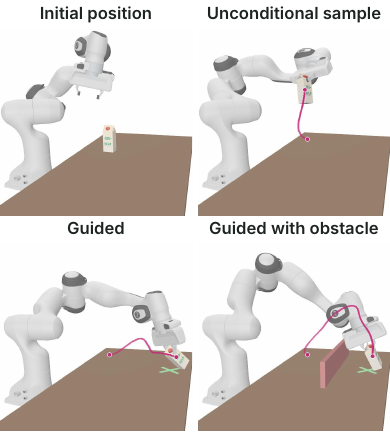}
  \relax
  \caption{\textbf{Planning with a generative interaction prior.}
  Latent noise optimization guides a frozen \model{} toward an object-space
  goal (green $\times$). When a wall blocks the direct path, the generated
  actuator trajectory lifts the object over it.}
  \label{fig:guidance}
  \relax
  \vspace{-2pt}
\end{wrapfigure}

 We compare action-
and noise-space planning on three contact-rich tasks, using task-specific
world models and rewards defined solely on desired object outcomes
(Fig.~\ref{fig:guidance}). All generated actuator trajectories are executed
\textbf{open-loop in simulation}, testing whether plans optimized through
the learned dynamics succeed under the underlying simulator.

\paragraph{Tasks and metrics.}
We evaluate three tasks requiring different contact strategies.
\textbf{Pushing} requires a rod to reposition around a rigid block before
pushing it toward a target over 60 steps.
\textbf{Pile Manipulation} requires a paddle to move a granular pile toward
a target over 150 steps, typically through multiple pushes.
\textbf{Picking} requires establishing a stable grasp and lifting an object
to a target over 60 steps. We evaluate five objects of different shapes
and sizes (milk carton, peanut-butter jar, soda can, pasta box, and water
bottle), each with five targets.

We report \emph{task progress}, defined as the fraction of the initial
object-to-target distance closed by the end of the episode. Performance is
measured across planning budgets of up to 100\,s for Pushing and 200\,s
for Pile Manipulation and Picking, on a single L40S GPU. Additional task
details are provided in Appendix~\ref{supp:planning_task_details}.

\paragraph{Planning methods.}
\emph{Noise-space planning.}
\textit{Latent Noise Optimization (LNO)} optimizes the initial Gaussian noise by backpropagating the terminal
reward through the generated trajectory. As a gradient-free alternative,
\emph{Noise CEM} applies the cross-entropy method in the same latent space:
each iteration samples candidate noise vectors, generates and evaluates
their trajectories, and refits the sampling distribution to the
highest-reward candidates. Both methods exploit the learned joint distribution
over actuator and object motion. Generated actuator trajectories are
executed using a simple tracking controller.

\emph{Action-space planning.}
\emph{Action CEM} instead optimizes raw per-frame end-effector
displacements, evaluating each candidate using an action-conditioned
world model. We apply Action CEM to both \model{} and
PointWorld~\citep{huang2026pointworld}. Crucially, action-space and
noise-space planning use the \emph{same \model{} checkpoint}, isolating
the effect of the search space rather than differences in model accuracy.
We vary planning budgets through the number of samples, optimization
iterations, or CEM rounds. Implementation details are provided in
Appendix~\ref{supp:planning_methods}.

\begin{figure}[t]
  \centering
  \includegraphics[width=\linewidth]{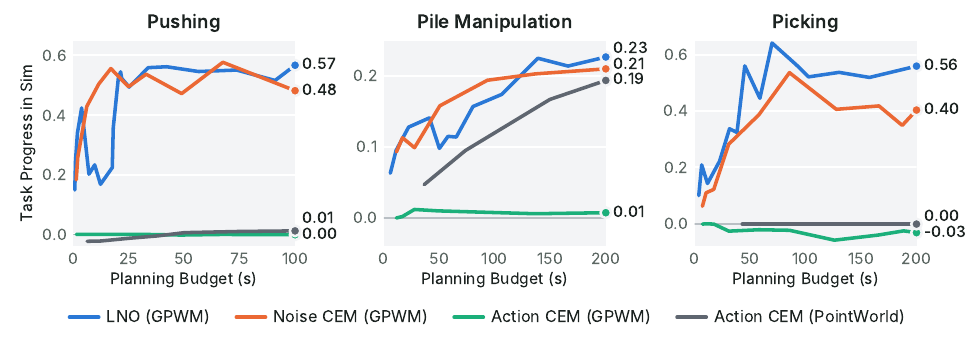}
  \caption{\textbf{Planning performance.}
  Task progress versus planning budget, measured after open-loop execution
  in simulation. Results are averaged over 50 scenarios for Pushing and
  Pile Manipulation and 25 problems for Picking. Reported values are
  interpolated at each task's planning budget (dashed lines).}
  \label{fig:planning_progress_figure}
\end{figure}

\paragraph{Generative interaction priors enable contact-rich planning.}
Figure~\ref{fig:planning_progress_figure} shows that searching \model{}'s
noise space substantially outperforms direct action-space optimization,
even when both use the same world model. On Pushing and Picking,
noise-space planners achieve 0.48--0.57 and 0.40--0.56 task progress,
respectively, while action-space planners remain at or below 0.01.
These tasks require discovering structured sequences of contact:
repositioning the rod before pushing, or establishing a grasp before
lifting. Such behaviors are unlikely to emerge from random action
proposals, whereas \model{}'s learned interaction prior provides
plausible trajectories that can be steered toward the goal.

Pile Manipulation presents a different regime: even random paddle
contacts can produce task progress, making action-space search more
competitive. Nevertheless, noise-space planning achieves 0.21--0.23
task progress, compared with 0.19 for the strongest action-space baseline.

Beyond discovering effective contacts, the generative prior also reduces
the risk of exploiting world-model errors. Direct action optimization can
select out-of-distribution trajectories that appear successful under the
learned dynamics but fail when executed in simulation, as illustrated
by Picking (Appendix~\ref{supp:planning_failures}). Noise-space optimization instead
searches through the model's learned distribution of interactions,
favoring actuator motions whose predicted effects are more likely to
transfer to the underlying simulator.

Together, these results demonstrate that \emph{accurate dynamics prediction
alone is insufficient for effective contact-rich planning}: the learned
distribution of actuator--object interactions provides a powerful
search prior, enabling object-space goals to be translated into actuator
trajectories without reference demonstrations or goal-specific policies.

Additional qualitative results, including baseline comparisons,
generalization to real-world images, and planning examples, are available
on our \href{https://robertren1122.github.io/gpwm/}{project page}.

\relax
\section{Limitations and Future Directions}
\relax

Several limitations remain. First, \model{} assumes access to complete,
temporally tracked 3D scene state. Our experiments therefore isolate physical
prediction from perception and do not address recovering persistent particles
and object identities from raw visual observations. An important next step is
to integrate 3D reconstruction and tracking, ultimately learning physical world
models directly from RGB or RGB-D video. 
Second, although a single Unified \model{} can model heterogeneous physical
regimes within one set of weights, it still trails domain-specific
models on some settings. This specialization gap suggests interference
between heterogeneous dynamics and motivates better data balancing, better-matched point density across domains,
domain-aware routing or mixture-of-experts architectures, and increased model
capacity as the diversity of physical experience grows. 
Finally, latent noise optimization requires differentiating through the full
generative process and, for long-horizon tasks, through multiple streaming
windows. Its computational cost therefore grows with planning horizon and
currently limits rapid online control. Faster generative parameterizations,
including flow maps~\citep{boffi2025buildconsistencymodellearning,
boffi2025flowmapmatchingstochastic}, could substantially reduce the cost
of long-horizon reward-guided planning, an exciting direction for future work.

\relax
\section{Conclusion}
\relax

We introduced \model{}, a generative 3D particle world-modeling framework for
physical prediction and contact-rich planning. Across rigid, deformable,
granular, and cloth dynamics, domain-specific \model{} models outperform prior
learned dynamics approaches, while a single Unified model demonstrates that
heterogeneous physical regimes can be modeled within one set of weights.
Flow matching with diffusion forcing enables stable auto-regressive generation far beyond
the training horizon, including coherent deformable-object prediction for more
than $55\times$ the trained prediction window. Beyond prediction, \model{} learns a generative prior over coupled
actuator--object motion that enables contact-rich planning from object-space
goals. Searching this learned interaction prior substantially outperforms
direct action-space optimization, even when using the same world model. Together, these results establish
generative particle world models as a common framework for long-horizon
physical prediction and goal-directed interaction in 3D.

\newpage

\bibliography{main}

\appendix
\clearpage
\section{\dataset{} Dataset Details}
\label{app:particleworlds}

\begin{figure}[!htbp]
  \centering
  \includegraphics[width=0.88\linewidth]{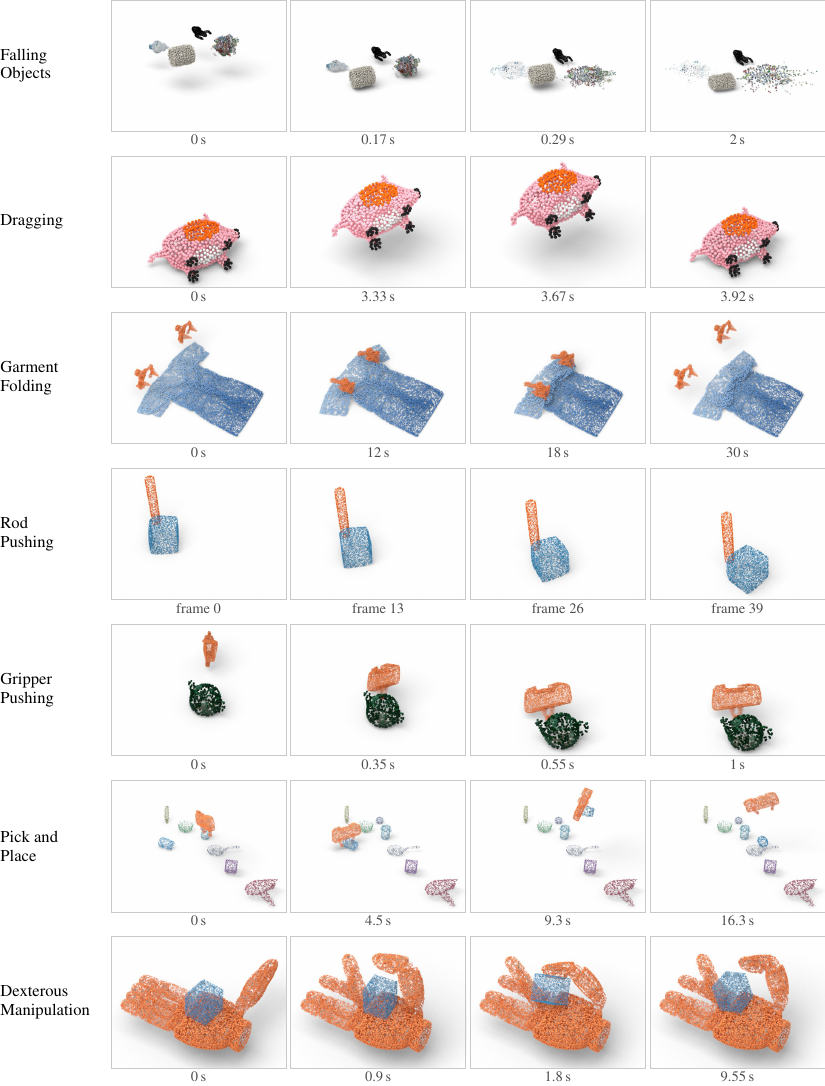}
  \caption{\textbf{Examples from the \dataset{} domains.} Each row shows one
  recorded episode at four moments, labeled with the time since its first
  frame; Pushing has one row for each of its two sources. From the top, four objects of three materials fall onto the table, a
  piggy bank is lifted by a patch on its back and released, two grippers fold a
  shirt, a rod pushes a block, a gripper pushes an object, a gripper picks up an
  object from a cluttered table and places it, and a hand turns a cube. Actuated
  parts are orange, and objects keep the colors of their meshes where
  available. Rigid bodies are drawn with extra points that follow their
  recorded poses, denser than the points the model receives. Rod Pushing
  records no time step, so its panels give frame indices.}
  \label{fig:app_dataset_examples}
\end{figure}

\begin{table}[!ht]
    \centering
    \caption{\textbf{Overview of \dataset{}.}
    Pushing combines rod and gripper recordings, and Pick and Place combines
    single-object and cluttered recordings. Frame
    counts refer to usable training episodes at each domain's training
    sampling rate, including recorded settling frames.}
    \label{tab:particleworlds}
    \vspace{0.03in}
    \small
    \setlength{\tabcolsep}{4pt}
    \renewcommand{\arraystretch}{1.08}
    \begin{tabular}{@{}lllr@{}}
        \toprule
        \textbf{Domain}
        & \textbf{Materials}
        & \textbf{Actuation}
        & \textbf{Training frames} \\
        \midrule
        Free Fall
        & Elastic, elastoplastic, granular
        & ---
        & 2{,}317{,}112 \\
        Dragging
        & Elastic, elastoplastic
        & Material-point velocities
        & 183{,}949 \\
        Garment Folding
        & Cloth
        & Two grippers
        & 4{,}068{,}020 \\
        Pushing
        & Rigid
        & Rod or gripper
        & 5{,}958{,}397 \\
        Pick and Place
        & Rigid
        & Gripper
        & 7{,}257{,}150 \\
        Dexterous Hands
        & Rigid
        & Articulated hand
        & 5{,}679{,}049 \\
        \midrule
        \multicolumn{3}{l}{\textbf{Total}}
        & \textbf{25{,}463{,}677} \\
        \bottomrule
    \end{tabular}
    \relax
\end{table}

\paragraph{Domains and counting convention.}
The six domains in Table~\ref{tab:particleworlds} group eight sources:
Pushing combines rod and gripper pushing, and Pick and Place combines
single-object and cluttered recordings.
The corpus contains 267{,}864 usable training episodes and 25{,}463{,}677
scene frames, counted once at each loader's temporal rate, including
recorded settling prefixes. Counts exclude test episodes and do not multiply
frames by overlapping windows, repeated training draws, or particle counts.
Rod and gripper pushing contribute 3{,}706{,}040 and 2{,}252{,}357 frames,
and single-object and cluttered Pick and Place contribute 1{,}690{,}120 and
5{,}567{,}030 frames, respectively.

\paragraph{Generation and source trajectories.}\mbox{}\\
\textbf{Free Fall and Dragging.} We generate both domains with the Genesis material-point
solver~\citep{Genesis}. Falling scenes contain one to four objects with
1{,}024 material points each and span elastic recovery, permanent deformation,
and granular rearrangement. Dragging uses elastic and elastoplastic objects:
prescribed velocities drive a persistent patch of material points through
pull, hold, and release events. Falling episodes contain 49 recorded states;
dragging episodes contain 49, 81, or 113 states, depending on the number of
events, with a recorded interval of 0.0417 seconds.

\textbf{Garment Folding.}
We generate scripted garment manipulation with two grippers in
SIM1~\citep{sim1}. After filtering, the domain covers 988 garments from ten
categories. Each clip contains approximately 900 states at 30\,Hz. The representation uses
a persistent farthest-point subset of the garment mesh with 8\,mm spacing,
capped at 10{,}000 cloth points, together with 1{,}000 points per gripper.
The same garment subset is used across frames and clips. These are
numerically screened manipulation trajectories, not verified successful folds.

\textbf{Rigid manipulation.}
Pushing has two sources. Rod Pushing uses our recordings of scripted
rod motion against a block,
with 40 states and 256 points per episode. Gripper Pushing uses our
reinforcement-learning policy recordings of successful object pushes.
Pick and Place recordings come from SkillWeaver~\citep{zhu2026skillweaver}.
Single-object Pick and Place combines recordings of four policies for
picking, picking convex objects, bringing objects upright, and placing held
objects. Its cluttered counterpart includes the target and surrounding
objects, which may move or remain still. For Dexterous Hands, we
generate trajectories from five teacher policies in Genesis using XHand1
and Sharpa hands for object rotation, reorientation, and screwdriver turning.
The rigid gripper and hand recordings are consumed at 20\,Hz.

\paragraph{Inputs, targets, and physical properties.}
Action-conditioned prediction receives scene history and the future motion
of rigid actuator links, and predicts passive-object dynamics. Cluttered
Pick and Place includes the surrounding objects in the prediction target.
Dragging instead supplies commanded material-point velocities, together
with indicators distinguishing a stationary hold from a release. These
commands need not equal the achieved motion: the target includes every
material point, including the driven patch. Passive falling scenes have no
action input. Recorded physical properties are supplied where available.
The MPM domain varies stiffness over three values while fixing Poisson's
ratio and density; garment parameters and the randomized mass and friction
of Gripper Pushing and Dexterous Hands are not recorded for conditioning.

\paragraph{Splits and coverage.}
The release holds out geometries for Free Fall, Dragging,
single-object Pick and Place, and Dexterous Hands, and splits
Garment Folding by garment identity. Cluttered Pick and Place uses held-out
scenes. Rod Pushing tests new
trajectories on the same geometry. Gripper Pushing holds out objects.

\paragraph{Quality control.}
Source-specific screening removes numerical failures and conspicuous
simulation artifacts. MPM checks cover unsettled starts, wall interactions,
collapse, and persistent fragmentation during dragging; cloth clips undergo
numerical screening. Rigid-object audits detect table penetration, abrupt
motion, excessive hand--object overlap, and other failures, with exclusion
manifests recording rejected episodes and reasons. We also check pose and
surface-template reconstruction against recorded particles. Screening reduces
artifacts but does not establish solver convergence.

\section{Model, Training, and Baseline Details}
\label{app:implementation}

\subsection{Architecture}
\label{app:architecture}
Every reported model uses the multi-resolution network of
Section~\ref{sec:architecture}, with the sizes in Table~\ref{tab:app_arch}.
The hierarchy has four levels. The first holds every particle. The second keeps,
per object, 256 anchors for elastic, plasticine, and sand objects, 128 for rigid
bodies and robot links, and 512 for cloth. The third and fourth levels divide
these counts by 4 and then by 2, so a Free Fall object of 1{,}024 particles is
represented by 256, 64, and 32 anchors. Downsampling aggregates the 8 nearest
tokens of each anchor within its object, and upsampling interpolates from the 3
nearest anchors. From the third level on, each block also carries one object's worth of
register tokens. Temporal attention is bidirectional within the window in every reported
model.

\begin{table}[!ht]
  \centering
  \small
  \caption{\textbf{Network sizes.} The six specialists share one configuration.
  Levels run from the particles (first) to the coarsest anchors (fourth).}
  \label{tab:app_arch}
  \begin{tabular}{@{}lcc@{}}
    \toprule
    & Specialist & Unified \\
    \midrule
    Parameters & 103.1\,M & 502.2\,M \\
    Anchors per object at the second level & \multicolumn{2}{c}{256 deformable, 128 rigid body or link, 512 cloth} \\
    Reduction to the third and fourth level & \multicolumn{2}{c}{4 and 2} \\
    Feature width per level & 64, 128, 256, 512 & 128, 320, 640, 1280 \\
    Encoder blocks per level & 1, 2, 4, 9 & 1, 2, 4, 8 \\
    Decoder blocks per level & 1, 2, 4 & 1, 2, 4 \\
    Attention head width & 64 & 64 \\
    Dropout & 0.1 & 0.1 \\
    \bottomrule
  \end{tabular}
\end{table}

\subsection{Training}
\label{app:training}
Each specialist is trained on one domain. The unified model is trained on
all of them, and every batch comes from one domain drawn with equal probability.
A training window has 4 context frames and 16 future frames. Particle displacements and rigid-body increments are normalized
per domain by the 99th percentile of their per-frame magnitude, separately
in the horizontal plane and along the vertical axis.

Future frames are noised with diffusion forcing. In 60\% of the windows every future frame draws its own noise level uniformly, and in the other 40\% the noise level rises linearly along the window, so nearer frames are less noisy than farther ones, as during streaming inference. Context frames stay clean, except
the history of passive objects receives Gaussian jitter with a standard
deviation of 0.05 in normalized units, which imitates the error of predicted
context during rollouts. The network predicts the clean trajectories, and the
particle and rigid-body terms of the loss are weighted equally ($\gamma=1$).

All \model{} models are trained with Muon \citep{jordan2024muon} on the weight matrices (learning rate $3\times10^{-4}$,
with the update scaled to match AdamW's RMS \citep{liu2025muon}) and AdamW on all remaining parameters
(learning rate $10^{-4}$, $\beta_1=0.9$, $\beta_2=0.999$), with decoupled weight decay
of $10^{-3}$ for both. Both learning rates follow a 5{,}000-step linear warmup and a
cosine decay to 1\% of their peak value. We clip gradients to a global norm of $1.0$,
keep an exponential moving average of the weights with decay $0.9999$, and train in
bf16 mixed precision. We train each specialist on two NVIDIA RTX PRO 6000 GPUs for 200k iterations,
and the unified model on four for 600k iterations.  Note that while the unified model has seen more training iterations, these update steps are split across 6 domains.  So the unified model gets only 100k iterations per domain, half the amount that the specialist models see.

\subsection{Baselines}
\label{app:baselines}
Each baseline is trained per domain on the same training episodes and
evaluated on the same test clips, with the same given frames and actuator
motion as our model unless noted.

\paragraph{PhysCtrl~\citep{wang2026physctrl}.}
We build on the authors' implementation and keep its spatiotemporal transformer and its diffusion objective. We scale it to about 100\,M parameters (8 blocks with hidden size 640) and train it from scratch on our data. The model receives the same history of 4 observed frames as ours, instead of the single frame it was designed for, predicts the same 16 frames per window, and processes every particle of the scene without subsampling.\\
PhysCtrl takes an action as regions of particles and the motion applied to each region. We cluster the actuator's particles by position into a few regions, with farthest point sampling choosing the region centers. The action is then given region by region. For every region and every predicted frame, the model receives two conditioning tokens, the position of the region's center and the region's average displacement in that frame. These tokens join the spatial attention of every frame, so every particle can attend to them. We do not give PhysCtrl rigid-body poses, since its interface does not support them. As for our model, the actuator's future motion is an input and is never predicted.\\
All objects share a single point cloud, and the spatial attention is all-to-all over the particles of each frame. Through PhysCtrl's
per-particle conditioning path, each particle is also tagged with the object it belongs to and that object's material parameters $(\log E, \nu, \log\rho)$.

  \paragraph{PointWorld~\citep{huang2026pointworld}.}
  PointWorld is a deterministic point-cloud dynamics model with a PTv3 backbone. We
  train it from scratch with its released code adapted to our setting: it
  operates on complete 3D point clouds and is applied recurrently to produce the
  autoregressive rollouts used in our evaluation. To match our 100\,M-parameter
  specialists, we use the released small variant with wider channels (about
  100\,M parameters), and train each run on the same episodes, splits, start frames
  and number of action-conditioned training windows as our corresponding
  specialist.\\
  PointWorld voxel-downsamples each point cloud before its backbone, by default on a
  $15$\,mm grid; on several of our domains this discards much of the objects'
  geometry, so we use a $1$\,mm grid instead, and at evaluation each surviving
  point's prediction is copied to the points merged into its voxel. We drop the
  input features our data does not provide, such as color and surface normals:
  each scene point carries only its positions over time and its distance to the
  robot, and each actuator point its positions, velocity and acceleration. We
  disable the training-time sphere crop, which removes scene points far from the
  actuator, so the model sees the whole scene, and fit its normalization
  statistics on $20{,}000$ training windows of each domain. The released optimizer
  and augmentations are otherwise unchanged.\\
  PointWorld receives the actuator's points at every frame and predicts only the
  scene, the actuator's future motion is an input and is never
  predicted. It observes a single context frame, where our models receive $4$, and
  we extend its prediction horizon from the released $10$ frames to $16$ to match
  our models. For cloth, training windows are drawn with the same motion weighting
  as our loader. Longer evaluation windows are rolled out autoregressively in
  $16$-frame chunks, each anchored on the model's own previous prediction.

  \paragraph{GNS~\citep{sanchez2020learning}.} We use GNS only for passive dynamics (Free Fall) where no actuator
  drives the scene; it is not run on the action-conditioned settings where PGND serves as the graph/grid-based
  baseline. We train it on the same episodes, splits, and start frames as our models, keeping the published
  architecture: $10$ message-passing steps, $128$-dimensional latents and a $0.015$\,m connectivity radius. It is a
  one-step model, so we roll it out autoregressively at every step to obtain a full-length prediction. GNS carries a
  learned embedding of a discrete particle type plus a material channel that is hard-coded to one scalar. We widen
  that channel to a per-particle vector and feed in each particle's normalized material properties.

  \paragraph{PGND~\citep{zhang2025particlegrid}.}
  We train PGND with $4$ given and $16$ predicted frames at each domain's native frame rate, rolling the model out
  one frame at a time with each prediction fed back as input. Its actuator representation needed the most change.
  PGND either drives material with a sphere collider or shows the network a rigid $500$-point gripper template posed
  by the end-effector, so its fingers never articulate. We instead feed the recorded actuator points every frame so
  finger motion is represented. The Dragging domain has no tool at all, since a patch of ${\sim}100$ material points is
  driven by a commanded velocity, so there we use PGND's sphere collider, placed on the patch at the start of each
  drag and then moved only by the commanded velocity, independent of the material. PGND does not condition on physics
  properties; on Dragging, where our model receives each object's material, we give PGND the same information as
  extra per-particle input channels. PGND simulates inside a fixed grid. Rather than excluding episodes that do not
  fit, we enlarge the grid or rescale the scene so that every training episode fits, and for cluttered Pick and Place
  we recenter a fixed $1.4 \times 1.0 \times 1.4$\,m grid on the gripper for each training window, freezing objects
  that leave it. In every domain the network's neighborhood radius covers the same $20$\,cm as PGND's released
  recipes. Otherwise, we train PGND with the same episodes, splits, and start frames as our own loader, and match each
  run to our corresponding specialist in gradient steps and windows per step.

\section{Evaluation Protocol}
\label{app:experiment_details}

\paragraph{Prediction protocol and metrics.}
All baseline models are domain-specific specialists, as are the
\model{} models in the Specialist row of
Table~\ref{tab:prediction_overview_proposed}. The Unified row uses a single
model trained jointly across all six \dataset{} domains. Methods are
evaluated on the same test trajectories with the same available actuator
conditioning. On action-conditioned tasks, recorded future actuator motion
is supplied throughout the rollout; passive-dynamics tasks require no action
conditioning. Predictions are fed back as context for subsequent windows.
For particle-tracked dynamics, we average Euclidean position error over
passive-object particles and predicted rollout frames, then average equally
across trajectories. Prescribed actuator motion is excluded from the error.
Cloth is evaluated using symmetric Chamfer distance, since particle
correspondence is not shared across all methods and surface geometry is the
quantity of interest.
Appendix~\ref{app:qualitative} shows predictions on every domain.

\paragraph{Test sets.}
Table~\ref{tab:app_eval_sets} lists the test clips behind each column. Every
rollout starts from 4 recorded frames and runs to the last recorded frame of its
clip, so horizons vary within a domain. A clip's error is its mean over the
predicted frames and the passive particles, and a column averages clips with
equal weight. Actuator particles, which are given, are not scored.

\begin{table}[!ht]
  \centering
  \small
  \setlength{\tabcolsep}{4pt}
  \caption{\textbf{Test clips.} The Held out column indicates how the test set differs from training.
  Predicted frames give the range of rollout lengths.}
  \label{tab:app_eval_sets}
  \begin{tabular}{@{}lcccc@{}}
    \toprule
    Column & Clips & Held out & Predicted frames & Frame interval \\
    \midrule
    Free Fall, 1 to 4 objects & 100 & shapes & 45 & 1/24\,s \\
    Free Fall, 5 to 7 objects (Table~\ref{tab:generalization_proposed}) & 100 & shapes and object count & 45 & 1/24\,s \\
    Dragging & 100 & shapes & 45 to 109 & 1/24\,s \\
    Pushing & 200 & objects & 16 to 93 & 0.05\,s \\
    Pick and Place & 200 & scenes & 233 to 921 & 0.05\,s \\
    Dexterous Hands & 50 & objects & 35 to 396 & 0.05\,s \\
    Garment Folding & 20 & garments & 898 & 1/30\,s \\
    \bottomrule
  \end{tabular}
\end{table}

\clearpage

\section{Additional Prediction Results}
\label{app:qualitative}

This section shows test clips for every domain of
Table~\ref{tab:prediction_overview_proposed}. Every model rolls out from the same four
recorded frames with the same actuator motion. The error under each panel is the mean position error of the
predicted particles at that frame.

\subsection{Free Fall}
\label{app:qual_freefall}

\begin{figure}[!ht]
  \centering
  \includegraphics[width=0.88\linewidth]{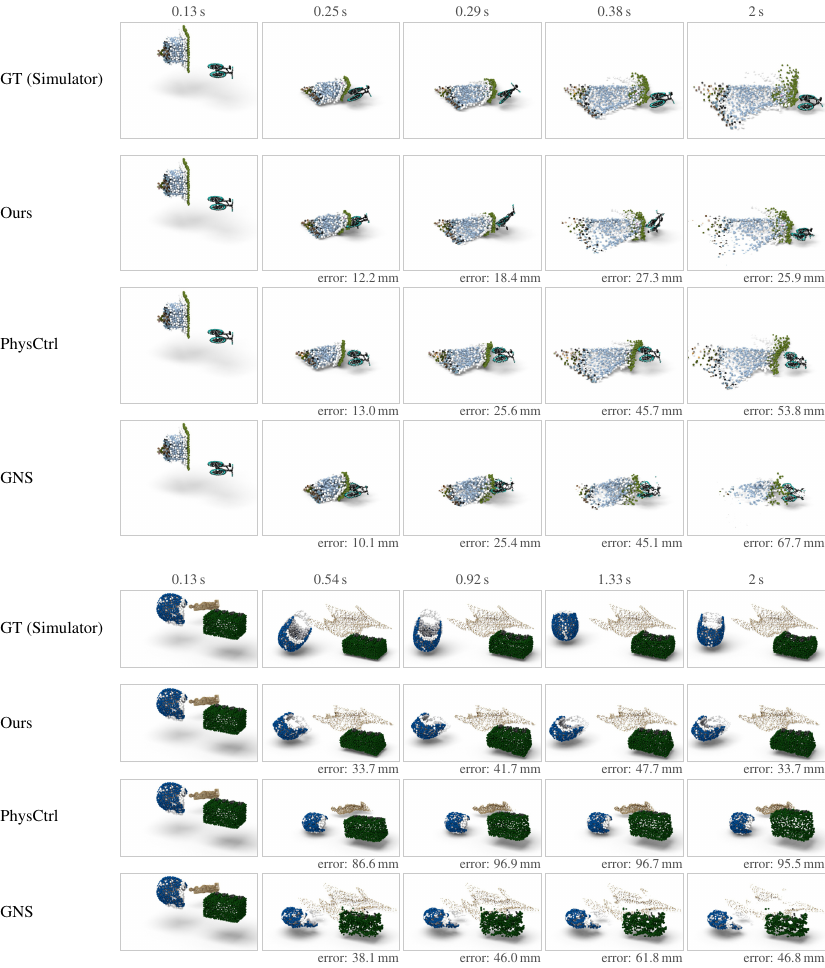}
  \caption{\textbf{Free Fall.} Two test clips with unseen shapes. Rows show the
  simulator's recording and three models, and columns show the last given frame,
  three moments of the fall and the last predicted frame, labeled with the time
  since release. Ours has the lowest clip error on 172 of the 200 Free Fall test
  clips, 84 of 100 with one to four objects and 88 of 100 with five to seven.}
  \label{fig:app_freefall}
\end{figure}

\begin{figure}[!ht]
  \centering
  \includegraphics[width=0.88\linewidth]{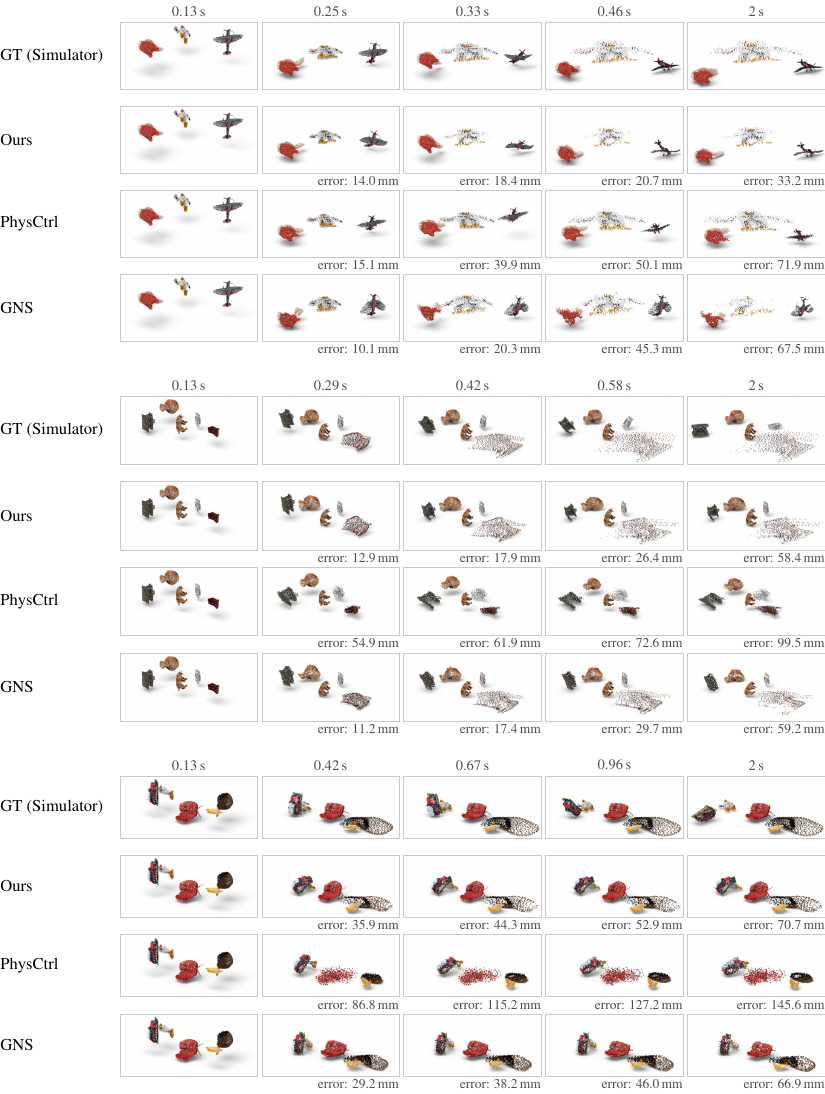}
  \caption{\textbf{Free Fall, continued.}}
  \label{fig:app_freefall_2}
\end{figure}

\clearpage
\subsection{Dragging}
\label{app:qual_dragging}

\begin{figure}[!ht]
  \centering
  \includegraphics[width=0.88\linewidth]{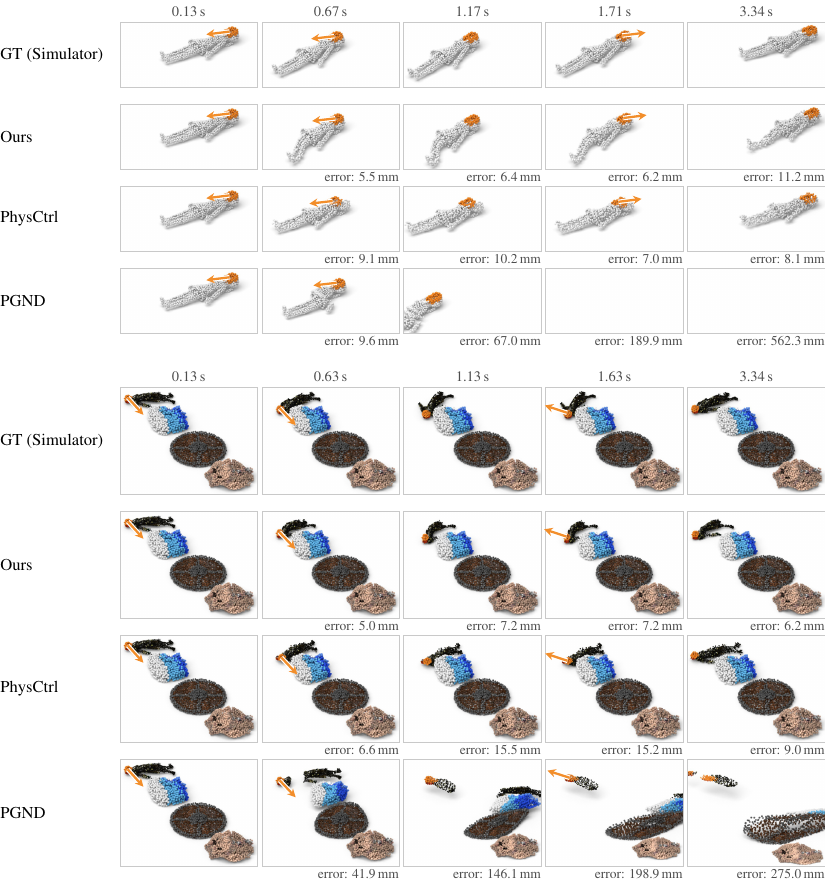}
  \caption{\textbf{Dragging.} Two test clips with unseen shapes, laid out as in
  Figure~\ref{fig:app_freefall}. The orange arrow shows the direction in which
  the dragged patch (orange) is pulled at that moment, from the recorded command
  that every model receives. Ours has the lowest clip error on 81 of the 100
  Dragging test clips.}
  \label{fig:app_dragging}
\end{figure}

\begin{figure}[!ht]
  \centering
  \includegraphics[width=0.88\linewidth]{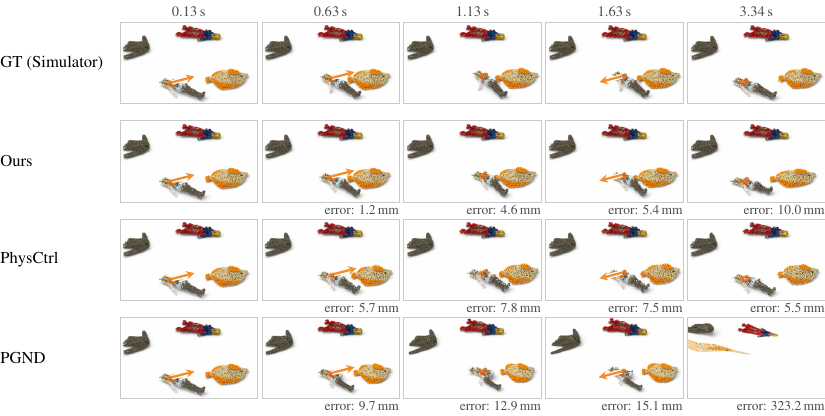}
  \caption{\textbf{Dragging, continued.} On this clip PhysCtrl has the lowest
  error.}
  \label{fig:app_dragging_2}
\end{figure}

\clearpage
\subsection{Garment Folding}
\label{app:qual_folding}

\begin{figure}[!ht]
  \centering
  \includegraphics[width=0.88\linewidth]{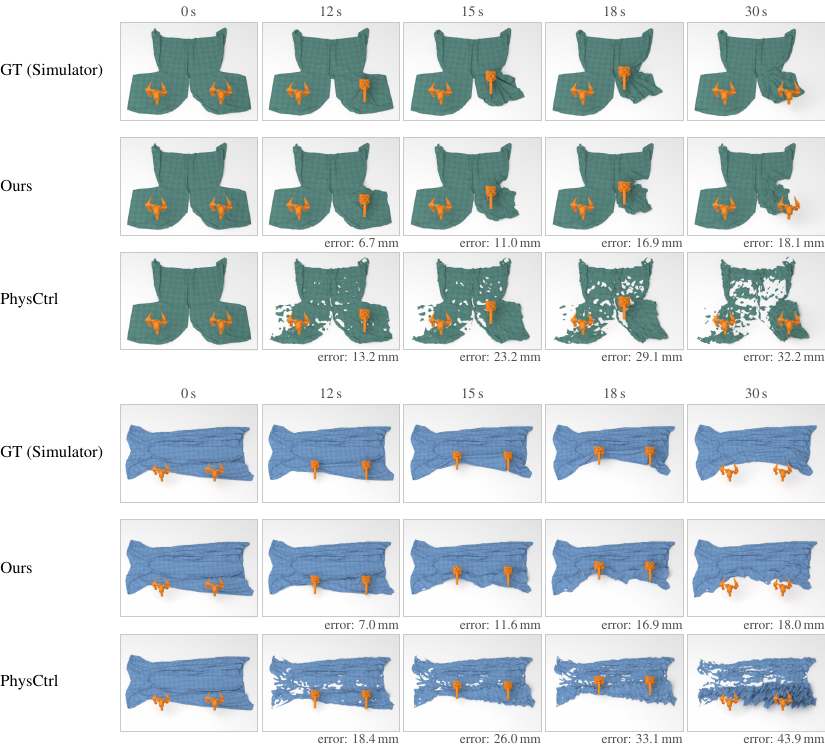}
  \caption{\textbf{Garment Folding, rendered back to the simulation mesh.} A jacket
  and a T-shirt not seen in training, 0 to 30\,s after the given history frames. Rows show
  the simulator and each model's particles driving the same mesh, with the grippers
  (orange) at their recorded poses. The error under each panel is the mean position
  error of the cloth particles at that moment. All rows are smoothed for better mesh visualization, which
  moves vertices by 1.5 to 6\,mm on average. Gaps in the PhysCtrl rows are where
  its cloth sinks below the floor.}
  \label{fig:app_folding}
\end{figure}

\begin{figure}[!ht]
  \centering
  \includegraphics[width=0.88\linewidth]{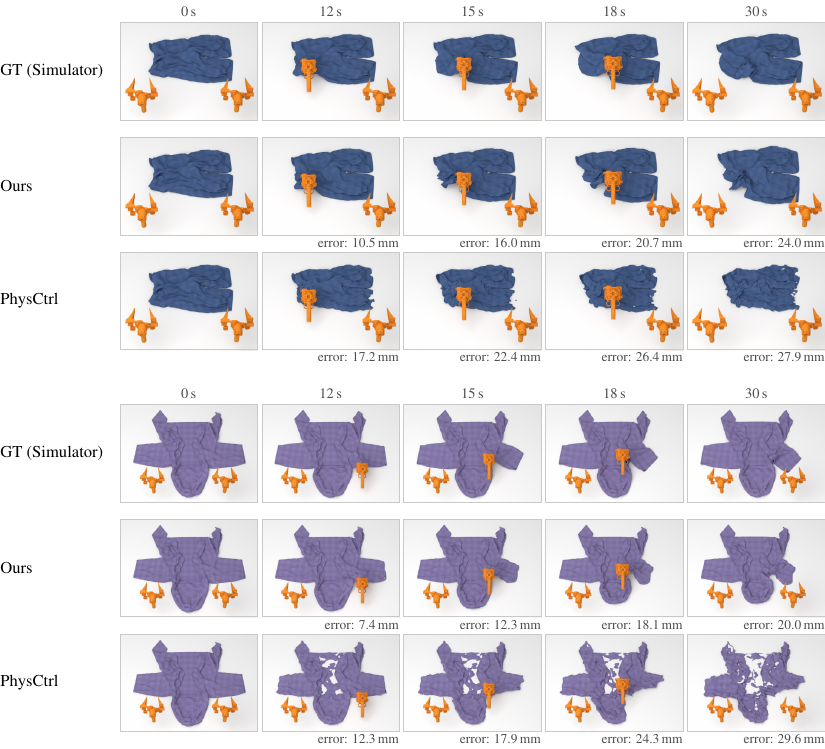}
  \caption{\textbf{Garment Folding, continued.} Pants and a hooded jacket.}
  \label{fig:app_folding_2}
\end{figure}

Figures~\ref{fig:app_folding} and~\ref{fig:app_folding_2} render predictions back
onto the simulator's own garment mesh. They use a model trained on the Garment
Folding domain of \dataset{} and four of its held-out test garments. 

\clearpage
\subsection{Pushing}
\label{app:qual_pushing}

\begin{figure}[!ht]
  \centering
  \includegraphics[width=0.88\linewidth]{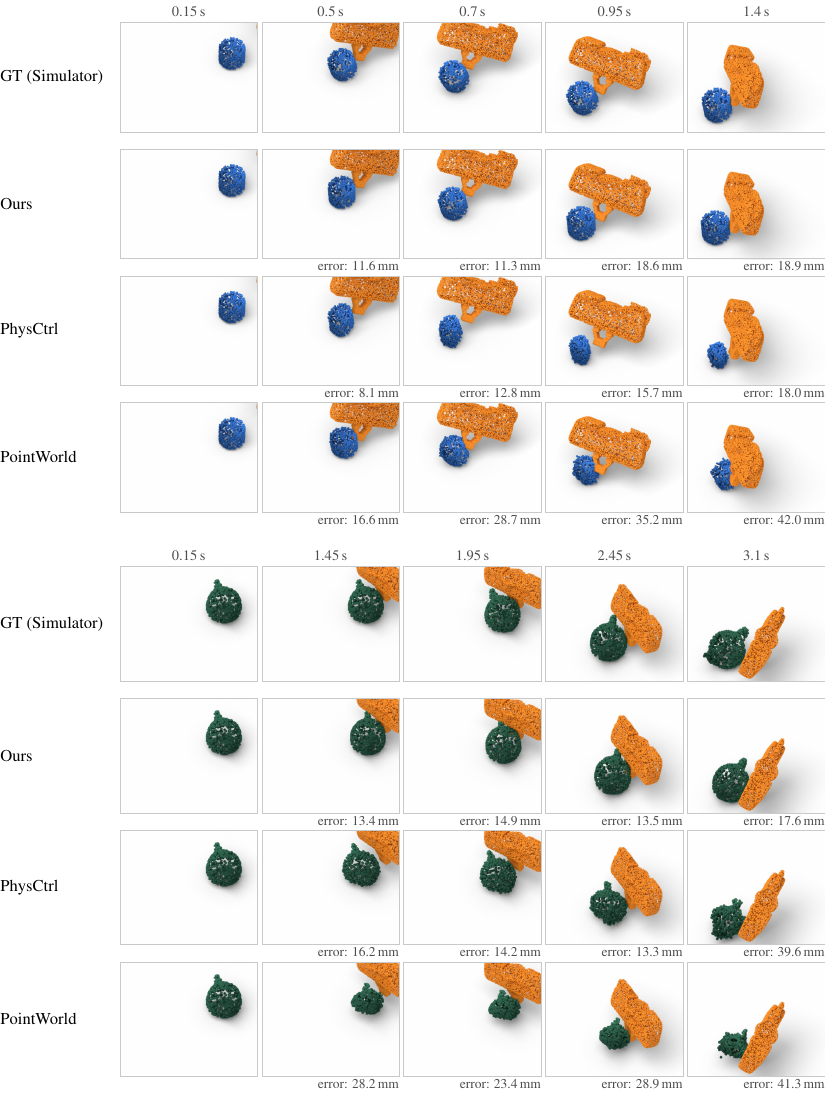}
  \caption{\textbf{Pushing.} Test clips with objects held out from training, laid
  out as in Figure~\ref{fig:app_freefall}. The gripper (orange) is given to every
  model and drawn at its recorded pose. Ours has the lowest clip error on 165 of
  the 200 Pushing test clips.}
  \label{fig:app_pushing}
\end{figure}

\begin{figure}[!ht]
  \centering
  \includegraphics[width=0.88\linewidth]{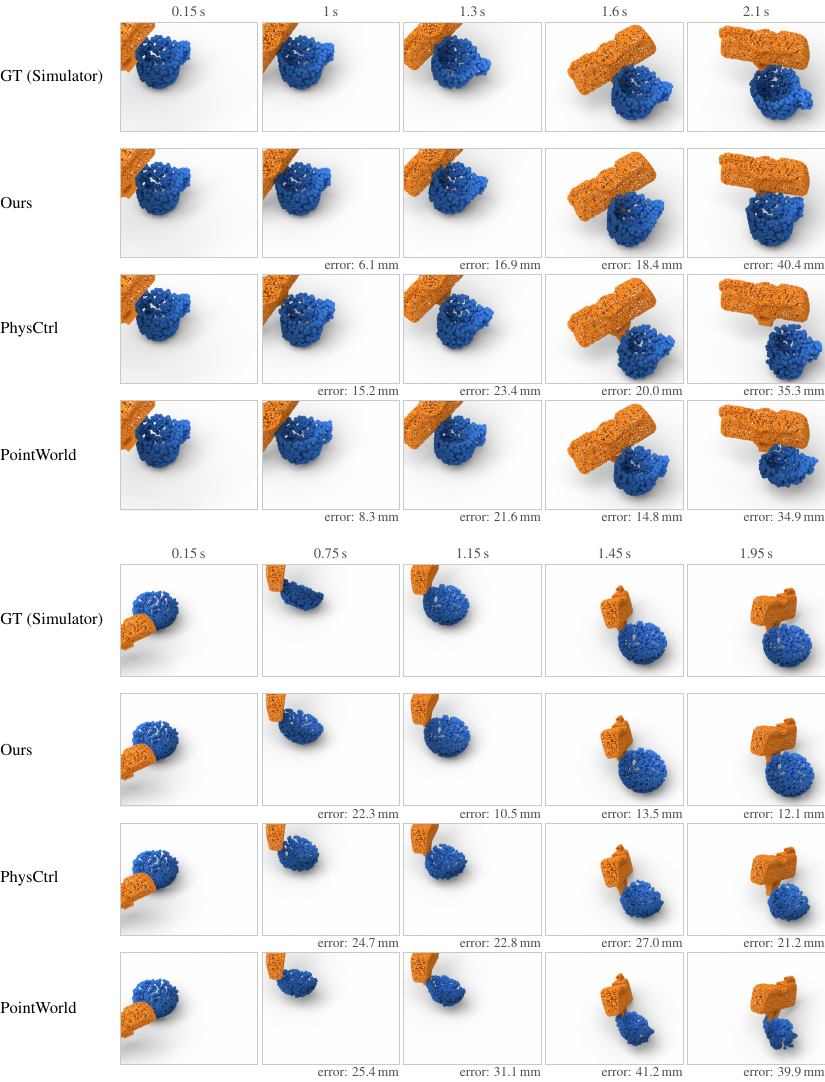}
  \caption{\textbf{Pushing, continued.}}
  \label{fig:app_pushing_2}
\end{figure}

\clearpage
\subsection{Pick and Place}
\label{app:qual_pick_place}

\begin{figure}[!ht]
  \centering
  \includegraphics[width=0.88\linewidth]{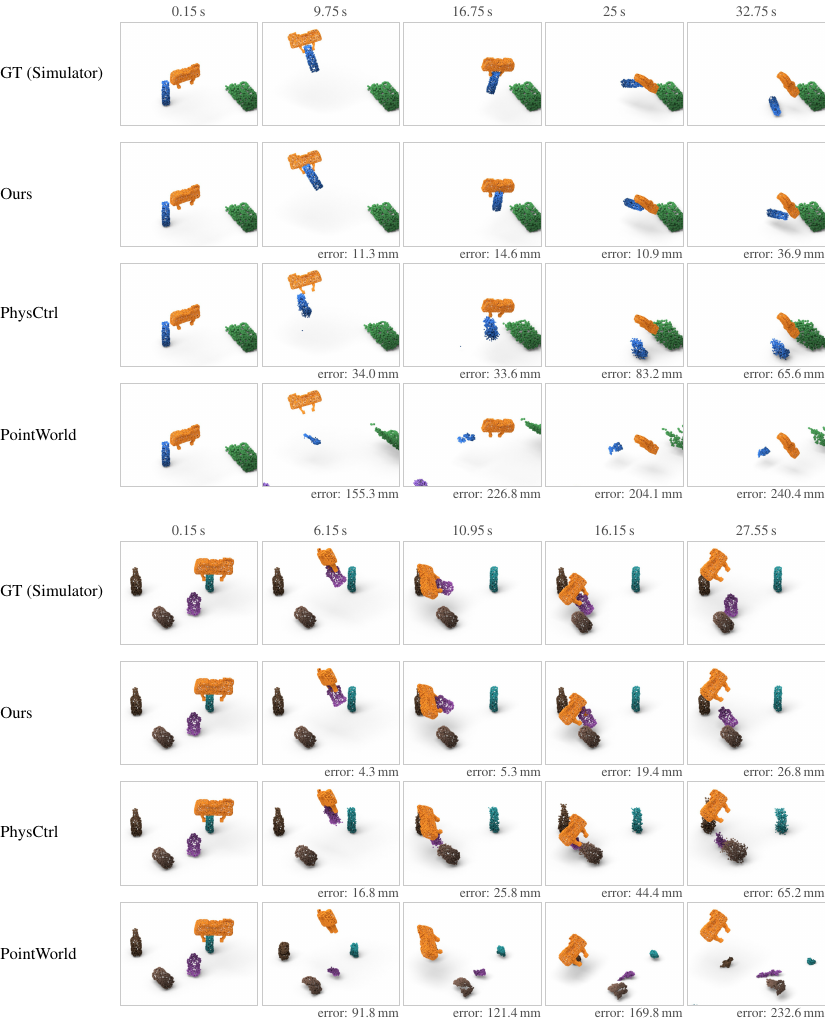}
  \caption{\textbf{Pick and Place.} Successful picks in held-out cluttered
  scenes, laid out as in Figure~\ref{fig:app_pushing}. Ours has the lowest clip
  error on 185 of the 200 Pick and Place test clips.}
  \label{fig:app_pick_place}
\end{figure}

\begin{figure}[!ht]
  \centering
  \includegraphics[width=0.88\linewidth]{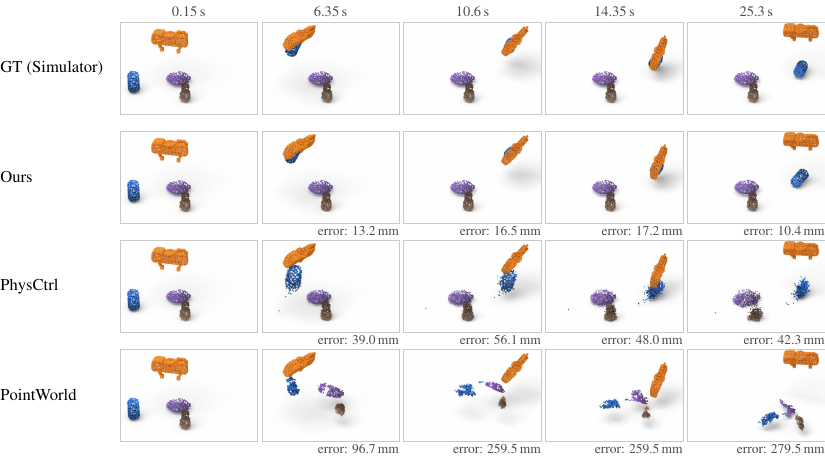}
  \caption{\textbf{Pick and Place, continued.}}
  \label{fig:app_pick_place_2}
\end{figure}

\clearpage
\subsection{Dexterous Hands}
\label{app:qual_dex}

\begin{figure}[!ht]
  \centering
  \includegraphics[width=0.88\linewidth]{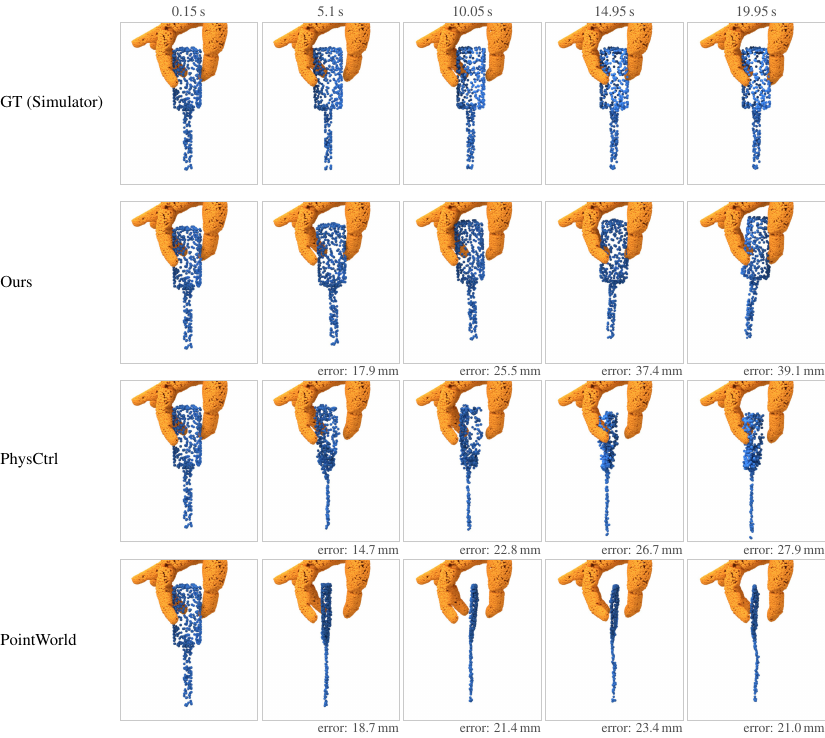}
  \caption{\textbf{Dexterous Hands.} A held-out screwdriver turned by the Sharpa
  hand. On this clip PhysCtrl and PointWorld have lower mean error than ours, but
  both move every point freely and thin the screwdriver.}
  \label{fig:app_dex}
\end{figure}

\clearpage
\section{Planning}
\label{app:planning}

\subsection{Planning Tasks}
\label{supp:planning_task_details}

In every task an actuator must bring a passive object to a target within
a fixed number of steps. The reward is sparse and defined only on the
final frame: the negative distance from the object's centroid to its
target. Each planner produces a single open-loop plan for the full
horizon inside its world model; the plan is then executed in the
task's simulator from the same start state, and the simulator never
enters planning.

\begin{figure}[!ht]
  \centering
  \includegraphics[width=\linewidth]{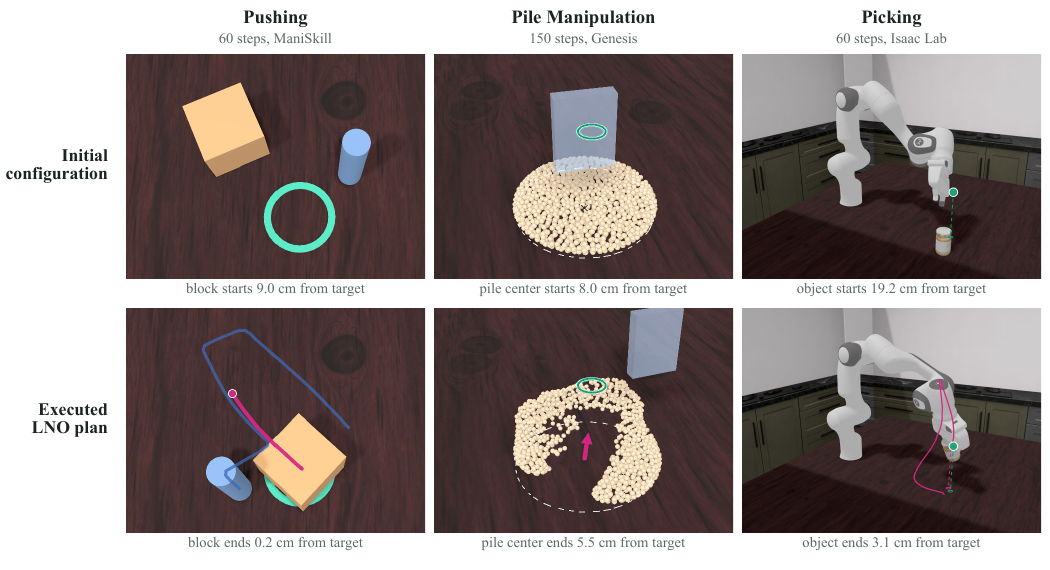}
  \caption{\textbf{Planning tasks.} One scenario per task: the initial
  configuration with its target (top) and the outcome of the LNO plan executed
  open-loop in the simulator (bottom). The rod moves around the block before
  pushing it; the paddle sweeps part of the pile onto the target; the gripper
  grasps the jar before carrying it to the target.}
  \label{fig:planning_tasks}
\end{figure}

\begin{itemize}[leftmargin=1em]
    \item \textbf{Pushing.} A rod pushes a rigid block toward a target on
    a tabletop. Targets typically require the rod to move around the
    block before pushing, so random rod motions rarely produce progress.
    Horizon 60 steps ($3\times$ the model window), 50 scenarios, executed
    in ManiSkill.
    \item \textbf{Pile Manipulation.} A paddle moves the centroid of a
    granular pile toward a target in the plane. World models are trained
    on $6{,}000$ trajectories of 300 frames from a scripted policy that
    randomly pushes and presses the pile. Horizon 150 steps
    ($7.5\times$ the model window), 50 scenarios, executed in Genesis.
    \item \textbf{Picking.} A parallel-jaw gripper must grasp a tabletop
    object (milk carton, peanut-butter jar, soda can, pasta box, or water
bottle) and carry it to a target 15\,cm above and up to 15\,cm to the
    side of its start (5 objects $\times$ 5 targets). Horizon 60 steps,
    executed with a floating Franka hand in Isaac Lab.
\end{itemize}
Figure~\ref{fig:planning_tasks} shows one scenario of each task and the
executed plan found by LNO.

\paragraph{Metric and budget.}
Task progress is $(d_0 - d_T)/d_0$, where $d_0$ and $d_T$ are the
object-to-target distances at the start and at the end of the executed
episode; it is averaged over scenarios. The planning budget is the
wall-clock time of the whole planning call on one L40S GPU. Each planner is run at a
ladder of settings (Table~\ref{tab:planning_ladders}), and the value
reported at a task's budget is linearly interpolated between the two
settings that bracket it.

\subsection{World Models}
\label{supp:planning_models}

\paragraph{\model{}.}
Each task uses a task-specific \model{} (100M parameters, 200k training
steps) trained on that task's data. During training, the actuator's
given motion is withheld in a random half of the windows and replaced by
a learned token, so the model learns both to generate the actuator
jointly with the scene and to follow a given actuator motion. Planning
uses the same checkpoint in both modes: noise-space planners use the
generative mode, and Action CEM uses the action-conditioned mode, in
which the actuator's motion is given as per-frame rigid displacements.
Rollouts over the full horizon use the streaming sampler of
Section~\ref{sec:inference} with 6 denoising steps.

\paragraph{PointWorld.}
Our PointWorld~\citep{huang2026pointworld} baseline is a deterministic,
recurrent PTv3 dynamics model, obtained by adapting PointWorld to operate on
complete 3D point clouds and to be applied recurrently
(Appendix~\ref{app:baselines}). It is trained on the same data, for a
number of action-conditioned training windows matched to \model{}'s,
conditions on the actuator's motion, and predicts the motion of the rest of
the scene; longer horizons are rolled out by chaining 16-frame
predictions, each starting from the previous prediction.

\subsection{Planners}
\label{supp:planning_methods}
\label{supp:planning_baselines}

\paragraph{Cross-entropy method.}
Noise CEM and both Action CEM variants use the same cross-entropy method
(\textbf{CEM}) and differ only in what it searches over. Each round draws
$K$ candidates from a Gaussian, rolls all of them out over the full
horizon, and refits the Gaussian to the best eighth of the candidates.
We use the elite-retention and variance-floor modifications of
iCEM~\citep{pinneri2021sample}. The best-scoring candidate found over all
rounds is the plan that is executed.

\paragraph{Noise CEM.}
The CEM variable is the noise of \model{}'s sampler. We use a stochastic
sampler, which draws fresh noise at every denoising step, so the searched
noise is the initial draw together with these per-step draws. With a
deterministic (ODE) sampler only the initial noise would remain, and
noise-space planning applies to either. Every candidate is a complete
generated trajectory, and the plan is its actuator motion.

\label{supp:noise_opt}
\begin{wrapfigure}{r}{0.45\textwidth}
    \centering
        \includegraphics[width=\linewidth]{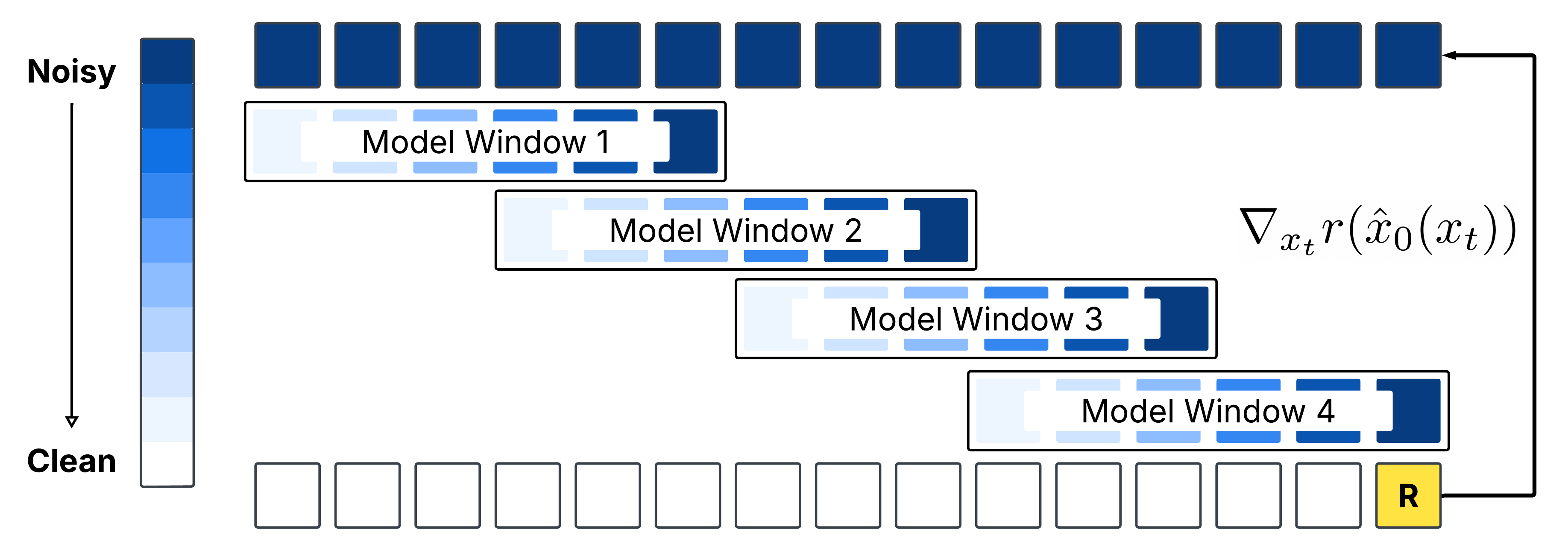}
        \caption{\textbf{Guidance step with streaming lookahead.}
        For a gradient step on noisy sample $x_t$, we get an estimate of the clean trajectory using streaming lookahead. Backpropagating through the lookahead allows guidance to optimize for long horizon tasks with sparse rewards only on the final frame.}
        \label{fig:guidance_step}
\end{wrapfigure}

\paragraph{Latent noise optimization.}
LNO optimizes the same noise by gradient ascent on the reward, with the
weights of \model{} frozen (Fig.~\ref{fig:guidance_step}). The streaming
sampler strides the model window across the full horizon of $T$ frames,
and the terminal reward
$r = -\lVert c(\hat{\mathbf{s}}_T) - g \rVert_2$ on the predicted object
centroid is backpropagated through every model call. LNO first draws a
pool of at least 32 random samples and keeps the best $P$; each is then refined
with $I=5$ AdamW steps (learning rate 0.1, gradient clipping at 1.0, and
a small penalty that keeps the noise statistically Gaussian). The member
with the highest predicted reward is executed.

\paragraph{Action CEM.}
Both world models are searched in the same action space: the actuator's
per-frame motion along the degrees of freedom it uses in the training
data.  This is the rod's planar translation for Pushing, the paddle's
translation and yaw for Pile Manipulation, and the hand's 6-DoF motion
plus gripper opening for Picking (the two fingers share one command).
Candidates are sampled independently per frame, and to keep motions in
distribution every step is clipped to the 99th percentile of the recorded
steps, both during planning and at execution.

\paragraph{Scaling the planning budget.}
A planner can spend more time in several ways: more candidates per CEM
round ($K$), more CEM rounds ($R$), more LNO members ($P$), or more
gradient steps per LNO member ($I$). For each planner we sweep the setting
that we found empirically to give the best performance for its planning
time, and keep the others fixed (Table~\ref{tab:planning_ladders}). Noise
CEM and Action CEM with \model{} sweep $K$, with $R$ fixed at three
rounds, so the two search spaces are compared at identical settings. LNO
sweeps $P$ with $I = 5$; for low planning budgets we also decrease the size of the initial screening pool. PointWorld sweeps $R$.

\begin{table}[!ht]
  \centering
  \caption{\textbf{Planner settings} swept by each curve in
  Fig.~\ref{fig:planning_progress_figure}; settings not listed are fixed.
  $K$: CEM candidates per round; $R$: CEM rounds; $P$: LNO members refined with
  5 gradient steps each. LNO's smallest budgets decrease the screening pool and take no gradient steps.}
  \label{tab:planning_ladders}
  \small
  \setlength{\tabcolsep}{5pt}
  \begin{tabular}{@{}llll@{}}
    \toprule
    Planner & \shortstack{Pushing\\(100\,s)} & \shortstack{Pile Manipulation\\(200\,s)} & \shortstack{Picking\\(200\,s)} \\
    \midrule
    LNO & pool 4--32; $P$ 1--48 & pool 4--32; $P$ 1--6 & pool 4--32; $P$ 1--12 \\
    Noise CEM & $K$ 4--192 & $K$ 4--64 & $K$ 4--96 \\
    Action CEM (\model{}) & $K$ 4--192 & $K$ 4--64 & $K$ 4--96 \\
    Action CEM (PointWorld) & $K{=}64$, $R$ 3--48 & $K{=}64$, $R$ 3--36 & $K{=}64$, $R$ 3--15 \\
    \bottomrule
  \end{tabular}
\end{table}

\subsection{Execution}
\label{supp:planning_execution}

For noise-space planning, the plan is the actuator motion that \model{}
generates. At every frame we fit a rigid pose to the predicted actuator
particles (Kabsch alignment to the actuator's initial geometry), and the
simulator tracks these poses open-loop: the rod and the paddle by pose
control, and the Franka hand as a floating base together with its gripper
opening. For action-space planning, with either world model, the planned
per-frame actions are executed directly.

\subsection{Additional Analysis of Planning Results}
\label{supp:planning_failures}

\begin{figure}[!ht]
  \centering
  \includegraphics[width=\linewidth]{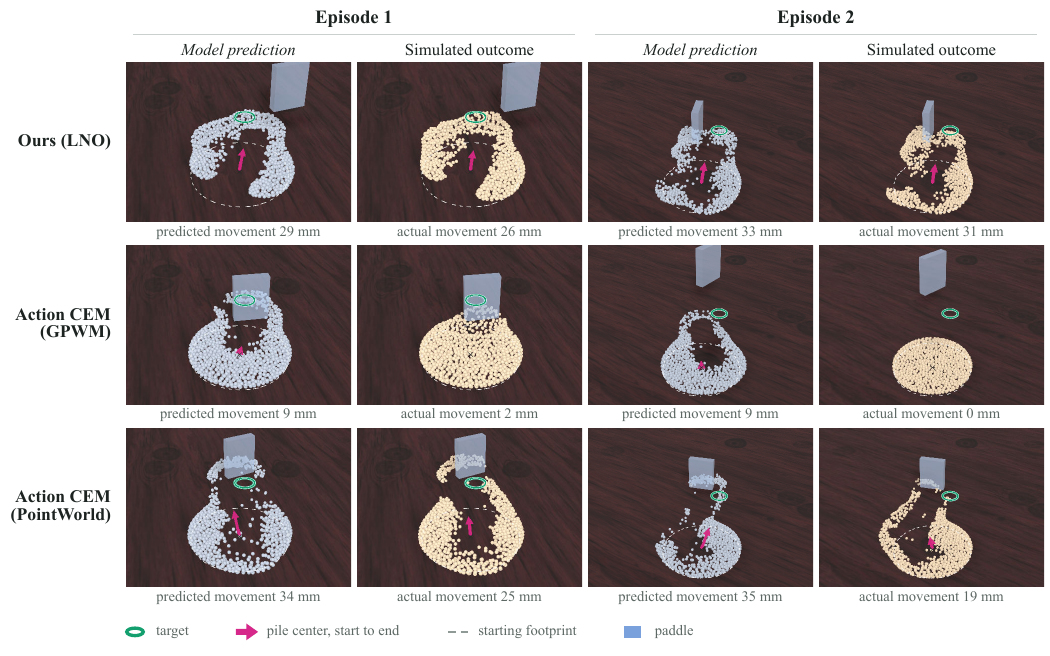}
  \caption{\textbf{Pile Manipulation.} Each planner's predicted outcome and the
  outcome of executing the same plan in Genesis, at the 200\,s budget. LNO's
  predicted push carries over to the simulator almost unchanged.}
  \label{fig:app_pile_manip}
\end{figure}

\paragraph{Predicted versus executed progress.}
Table~\ref{tab:planning_imagined} compares each planner's progress as
predicted by its world model with the progress it achieves in
simulation. Noise-space plans lose part of their predicted progress on
Pushing and Picking but keep most of it, and on Pile Manipulation their
predictions carry over almost exactly. Action-space plans predict
progress on every task and achieve almost none, with the exception of
PointWorld on Pile Manipulation. Figure~\ref{fig:app_pile_manip} shows each
planner's predicted and simulated outcomes on Pile Manipulation.

\begin{table}[!ht]
  \centering
  \caption{\textbf{Predicted and executed task progress} (\%), at each
  task's planning budget: progress in the world model's own rollout of the
  chosen plan $\rightarrow$ progress after open-loop execution in
  simulation.}
  \label{tab:planning_imagined}
  \small
  \setlength{\tabcolsep}{6pt}
  \begin{tabular}{@{}lccc@{}}
    \toprule
    Planner & Pushing & Pile Manipulation & Picking \\
    \midrule
    LNO & 89 $\rightarrow$ 57 & 27 $\rightarrow$ 23 & 89 $\rightarrow$ 56 \\
    Noise CEM & 94 $\rightarrow$ 48 & 21 $\rightarrow$ 21 & 70 $\rightarrow$ 40 \\
    Action CEM (\model{}) & 23 $\rightarrow$ 0 & 7 $\rightarrow$ 1 & 78 $\rightarrow$ 0 \\
    Action CEM (PointWorld) & 37 $\rightarrow$ 1 & 29 $\rightarrow$ 19 & 98 $\rightarrow$ 0 \\
    \bottomrule
  \end{tabular}
\end{table}

\begin{figure}[!ht]
  \centering
  \includegraphics[width=\linewidth]{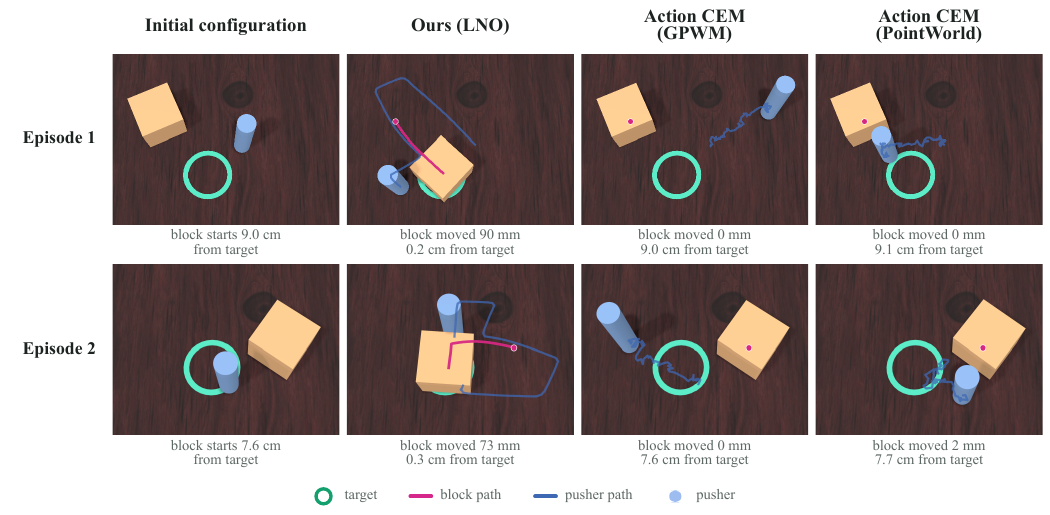}
  \caption{\textbf{Pushing.} Executed plans at the 100\,s budget on two
  scenarios. LNO in \model{} moves the rod around the block and pushes it onto
  the target. Action CEM, with \model{} or PointWorld, produces jittering rod
  motions that leave the block where it is.}
  \label{fig:app_block_push}
\end{figure}

\begin{figure}[!ht]
  \centering
  \includegraphics[width=\linewidth]{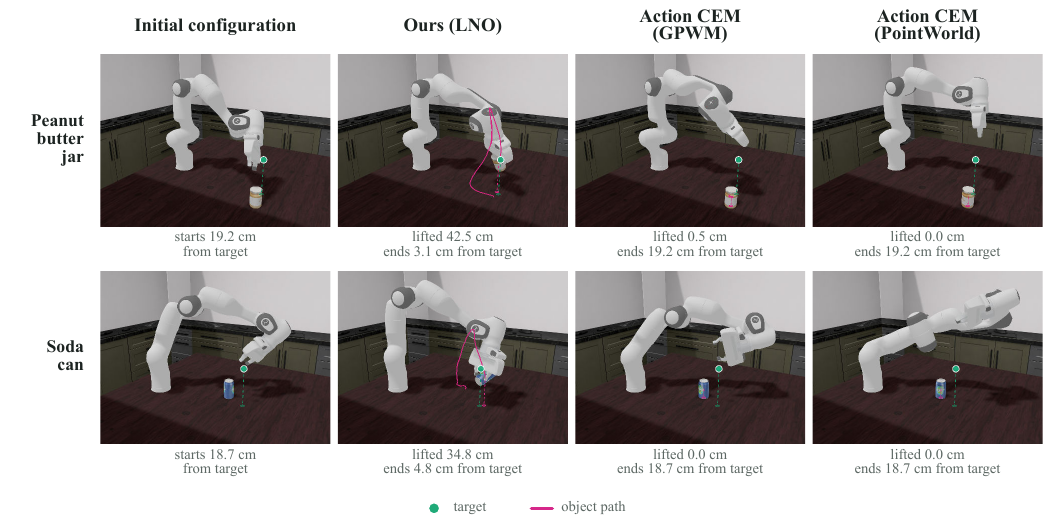}
  \caption{\textbf{Picking.} Executed plans at the 200\,s budget for two
  objects. LNO in \model{} grasps and lifts each object to the target. Neither
  Action CEM plan establishes a grasp, and the objects are not lifted. The full
  Franka arm is rendered for visualization only; plans are executed with a
  floating hand.}
  \label{fig:app_picking}
\end{figure}

\begin{figure}[!ht]
  \centering
  \includegraphics[width=\linewidth]{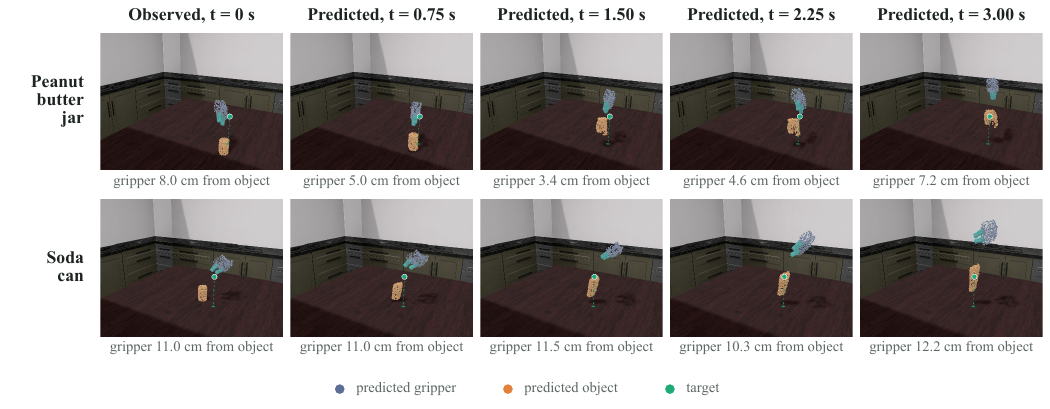}
  \caption{\textbf{PointWorld's predictions for its Picking plans.} The
  predicted object rises to the target while the predicted gripper stays
  3--12\,cm away from it; executed in simulation, the same plans leave the
  object on the table (Fig.~\ref{fig:app_picking}, right).}
  \label{fig:app_pointworld_pick}
\end{figure}

\paragraph{Pushing and Picking.}
Action-space search in either world model does not find the contact
sequence these tasks require, and Table~\ref{tab:planning_imagined} shows
that it fails differently on the two tasks. On Pushing, the search fails
to find progress even inside the world models: its chosen plans are
predicted to make only 23--37\% progress, against 89--94\% for
noise-space plans. Under a sparse terminal reward, random per-frame
motions that bump the block without pushing it toward the target give the
search little to improve on, and the rod rarely moves around the block.
On Picking, the search does find plans the world models rate highly
(78--98\% predicted progress), but none of that progress occurs in
simulation. No progress is possible before a grasp, and a grasp is hard
to discover in the high-dimensional space of hand and gripper motions.
The search instead settles on plans whose predicted object motion comes
from exploiting errors in the predicted dynamics.
Noise-space plans in \model{} move the rod behind the block before
pushing and grasp before lifting; these contact sequences come from the
learned interaction prior, not from raw search over actions
(Figs.~\ref{fig:app_block_push} and~\ref{fig:app_picking}).
PointWorld's predictions for its chosen Picking plans show the object
rising to the target while the gripper stays 3--12\,cm away from it
(Fig.~\ref{fig:app_pointworld_pick}).

\clearpage

\section{Predictions from a Single Image}
\label{app:real_photo}

This section gives the details of the predictions from single images in
Section~\ref{sec:exp_generalization_proposed} and more examples.

\subsection{Reconstruction and Setup}
\label{app:real_photo_setup}

\paragraph{Reconstruction.}
SAM~3~\citep{sam3} segments each object named in a text prompt,
SAM~3D~\citep{sam3d} reconstructs it as a mesh, and MoGe-2~\citep{moge2}
estimates the metric depth and camera that place the meshes in the scene.
Each object is represented by 1{,}024 particles sampled on its surface, as in
the training data.

\paragraph{Scale, materials, and release.}
Each scene is rescaled so that its objects match the particle spacing of the training
objects, by 1.7 times for the desk photograph, and times are reported at the
original scale. Each object is assigned a material type and a Young's
modulus. The four
given frames describe free fall from the chosen release pose.

\paragraph{Rendering.}
Image pixels of moving objects are filled with
LaMa~\citep{suvorov2022resolution}. Each object mesh follows its predicted
particles, every vertex moving with its four nearest particles, and is path
traced with the camera of the image. Sand is drawn as grains, and sand
particles that end below the table are drawn on its surface.

\subsection{More Examples}
\label{app:real_photo_more}
\begin{figure}[!htbp]
  \centering
  \includegraphics[width=0.86\linewidth]{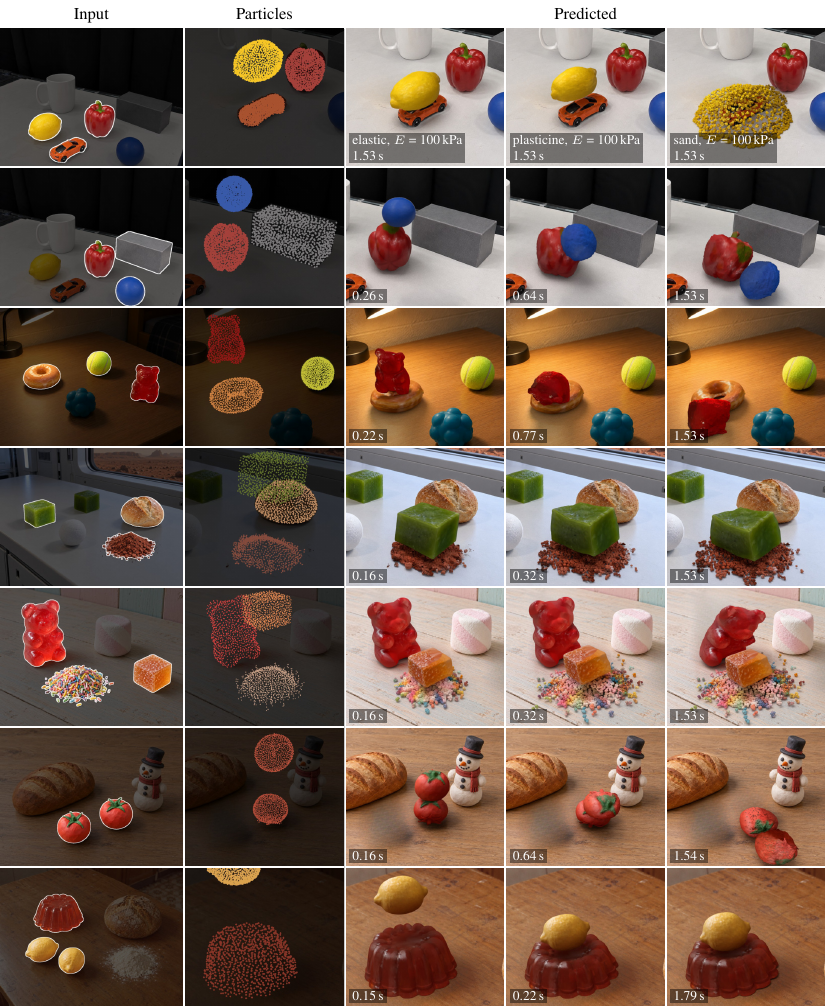}
  \caption{\textbf{More drops from images.} Each row shows the input image with
  the simulated objects outlined, their particles at the last input frame, and
  three predicted frames. Row 1, the lemon from the desk photograph falls onto
  the toy car as an elastic, a plasticine, and a sand object, shown 1.53\,s after
  release. Row 2, a plasticine ball lands on the pepper, rolls off and pushes it
  aside. Rows 3 to 7 use generated images. Row 3, an elastic gummy bear falls
  onto a plasticine donut on a dorm desk, flattens and slides off. It stays
  flattened, a limitation of the model for soft elastic objects. Row 4, an
  elastic gel cube falls onto a pile of Mars regolith simulated as sand and
  scatters it. Row 5, an elastic jelly cube falls onto sprinkles simulated as
  sand, and the gummy bear beside it slumps by 1.53\,s, the same limitation. Row 6, in a claymation scene, a plasticine
  tomato lands on an elastic one, and the two topple and come to rest apart.
  Row 7, in a 1970s photograph, an elastic lemon lands on an elastic jelly mold
  and comes to rest on top. Rows 6 and 7 are stylized images beyond the
  training data. Young's moduli are 10\,kPa for the gummy bears, donut, gel
  cube, bread roll, jelly cube, and jelly mold, 1\,MPa for the toy car and the
  block, and 100\,kPa for all other objects. The car, pepper, block, and tennis
  ball are elastic. Times are at each scene's original scale.}
  \label{fig:app_more_drops}
\end{figure}

Figure~\ref{fig:app_more_drops} shows more drops from the desk photograph and
from five generated images, two of them stylized images beyond the training
data.

\subsection{Changing the Observed History}
\label{app:nudge}
\begin{figure}[!ht]
  \centering
  \includegraphics[width=\textwidth]{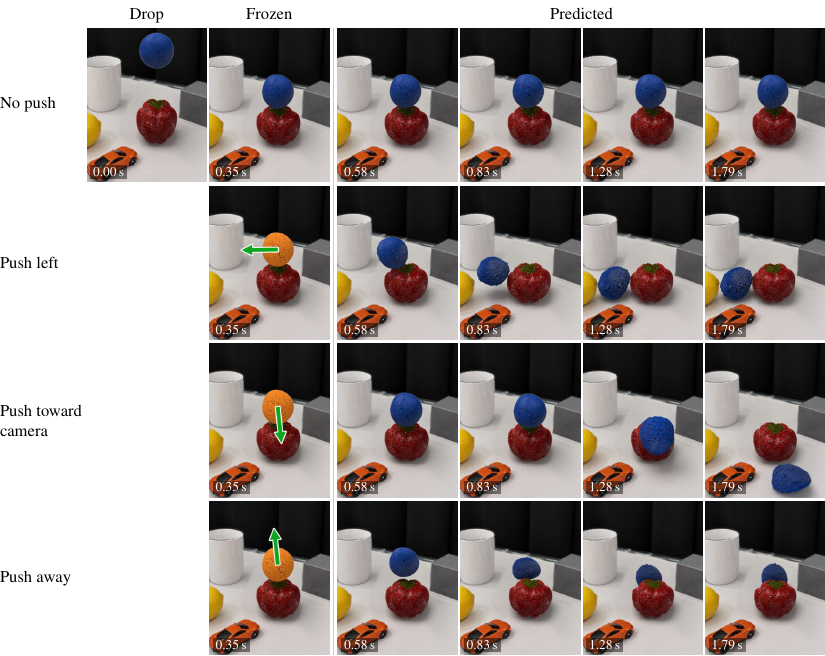}
  \caption{\textbf{Changing the observed history.} Every row continues the same
  drop of the blue ball onto the pepper. In the top row the ball is released,
  lands, and rests on top at 0.35\,s, when we freeze the scene. Without an edit
  it stays there. In each row below, only the ball's four input frames at the
  freeze are rewritten so that it moves in the direction of the green arrow (the
  edited ball is drawn orange), and the model predicts the rest. Times are since
  the release at the original scale. The last column is 1.44\,s after the
  freeze.}
  \label{fig:nudge}
\end{figure}

The model receives four frames of history, so editing them changes the
predicted future. In Figure~\ref{fig:nudge} a ball dropped onto a pepper comes
to rest on top. We then rewrite only the ball's four input frames so that it
moves sideways at 9\,cm/s from where it rests, and let the model continue. No
actuator is modeled.

\subsection{Dragging a Reconstructed Object}
\label{app:sloth}
\begin{figure}[!ht]
  \centering
  \includegraphics[width=\textwidth]{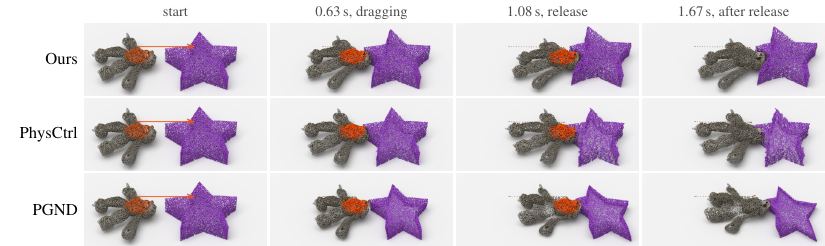}
  \caption{\textbf{Dragging a sloth into a foam star.} A sloth reconstructed from a photograph is dragged into a foam star, and every model gets the same start and drag. The arrow shows the commanded drag. Objects are drawn with dense points that follow each model's predicted particles.}
  \label{fig:sloth_drag}
\end{figure}

In Figure~\ref{fig:sloth_drag}, a sloth reconstructed from a photograph is
dragged into a foam star. Every model receives the same initial scene and
commanded drag, and the arrow shows the drag direction.

\subsection{Folding Garments from Generated Images}
\label{app:garments}
\begin{figure}[!ht]
  \centering
  \includegraphics[width=\textwidth]{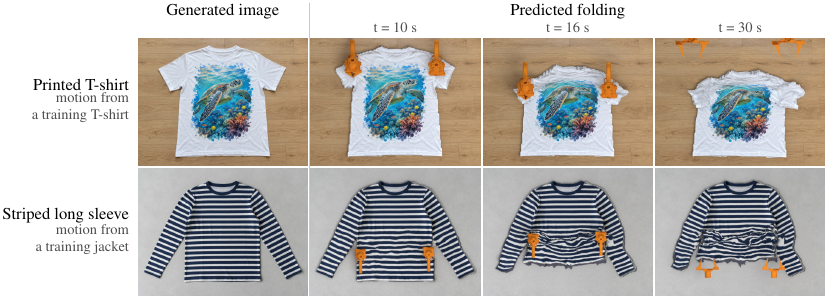}
  \caption{\textbf{Folding garments from generated images.} The grippers
  (orange) follow the motion recorded for a training garment, and their points
  are given to the model.}
  \label{fig:fold_from_image}
\end{figure}

Figure~\ref{fig:fold_from_image} applies the Garment Folding specialist to
garments from generated images. Each garment is segmented from a top-down
image, represented as two cloth layers at the particle spacing of the training
data, and placed on the most similar training garment, which sets its size and
supplies the recorded gripper motion. The specialist predicts all frames
from that motion, which moves the grasped edge 16 to 20\,cm, so each garment
ends in a partial fold.

\end{document}